\documentclass{article}

\usepackage{microtype}
\usepackage{graphicx}
\usepackage{subcaption}
\usepackage{booktabs} 

\usepackage{hyperref}

\usepackage[preprint]{icml2026}

\usepackage{amsmath}
\usepackage{amssymb}
\usepackage{mathtools}
\usepackage{amsthm}
\usepackage{bbding}
\usepackage{multirow}
\usepackage{multicol}
\usepackage{array}
\definecolor{myyel}{rgb}{.99,.92,.41}
\definecolor{mycy}{rgb}{.99,0.5,0.6}
\usepackage[capitalize,noabbrev]{cleveref}

\theoremstyle{plain}

\theoremstyle{definition}

\theoremstyle{remark}

\usepackage[textsize=tiny]{todonotes}

\icmltitlerunning{3DGSNav}

\begin{document}

\twocolumn[
  \icmltitle{3DGSNav: Enhancing Vision-Language Model Reasoning for \\
  Object Navigation via Active 3D Gaussian Splatting}



  \icmlsetsymbol{equal}{*}

  \begin{icmlauthorlist}
    \icmlauthor{Wancai Zheng}{1}
    \icmlauthor{Hao Chen}{2}
    \icmlauthor{Xianlong Lu}{1}
    \icmlauthor{Linlin Ou}{1}
    \icmlauthor{Xinyi Yu}{1}
  \end{icmlauthorlist}

  \icmlaffiliation{1}{Zhejiang University of
		Technology, Hangzhou, China.}
  \icmlaffiliation{2}{Zhejiang University, Hangzhou, China}
  \icmlcorrespondingauthor{Xinyi Yu}{yuxy@zjut.edu.cn}

  \icmlkeywords{Machine Learning, ICML}

  \vskip 0.3in
]



\printAffiliationsAndNotice{}  
\begin{abstract}
Object navigation is a core capability of embodied intelligence, enabling an agent to locate target objects in unknown environments. Recent advances in vision–language models (VLMs) have facilitated zero-shot object navigation (ZSON). However, existing methods often rely on scene abstractions that convert environments into semantic maps or textual representations, causing high‑level decision making to be constrained by the accuracy of low‑level perception. In this work, we present 3DGSNav, a novel ZSON framework that embeds 3D Gaussian Splatting (3DGS) as persistent memory for VLMs to enhance spatial reasoning. Through active perception, 3DGSNav incrementally constructs a 3DGS representation of the environment, enabling trajectory-guided free-viewpoint rendering of frontier-aware first-person views. Moreover, we design structured visual prompts and integrate them with Chain-of-Thought (CoT) prompting to further improve VLM reasoning. During navigation, a real‑time object detector filters potential targets, while VLM‑driven active viewpoint switching performs target re‑verification, ensuring efficient and reliable recognition.
Extensive evaluations across multiple benchmarks and real‑world experiments on a quadruped robot demonstrate that our method achieves robust and competitive performance against state‑of‑the‑art approaches.
\href{https://aczheng-cai.github.io/3dgsnav.github.io/}{{The Project Page:}} \text{https://aczheng-cai.github.io/3dgsnav.github.io/}

\end{abstract}

\begin{figure}[t]
    \centering
    \includegraphics[scale=0.32]{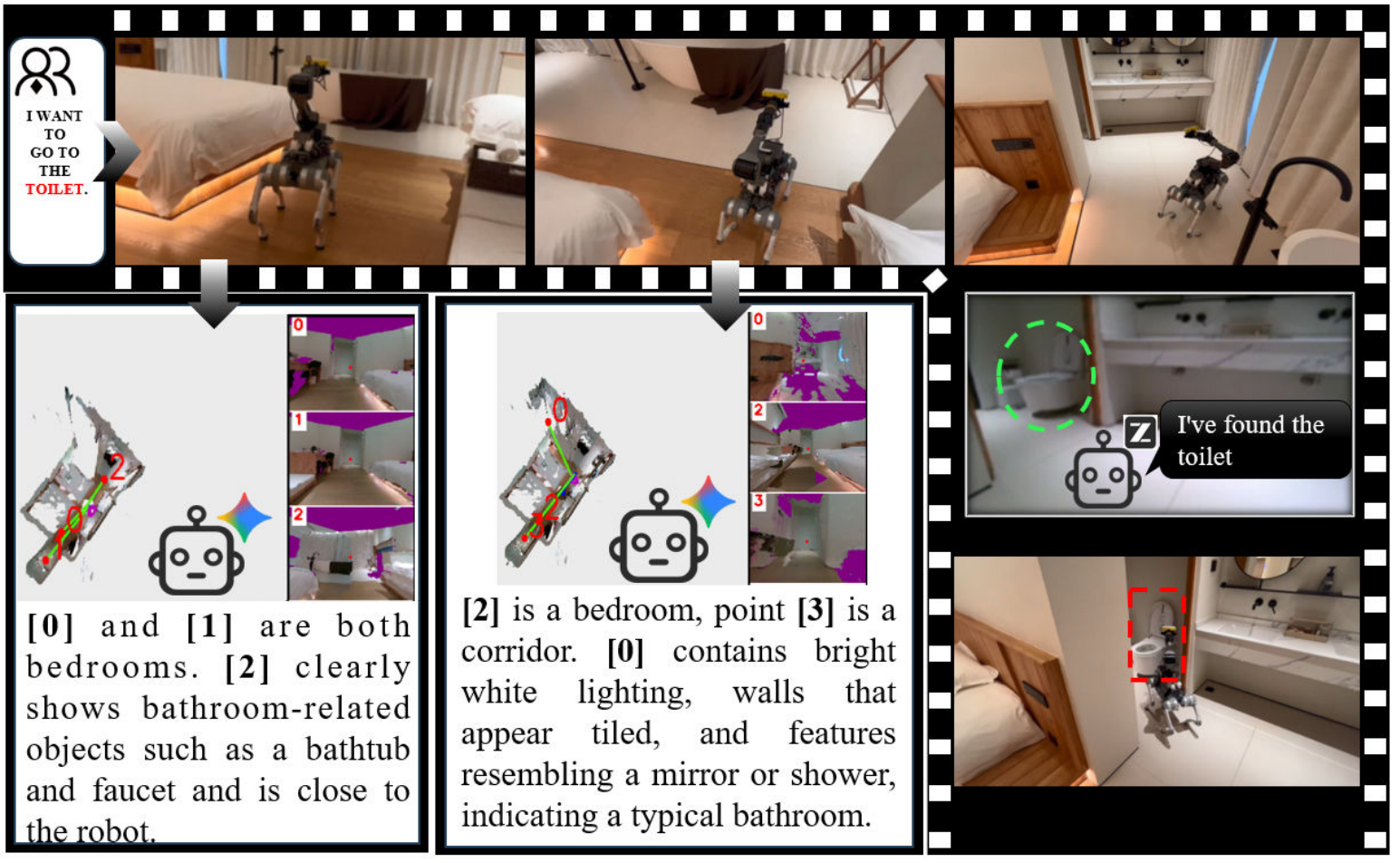}
    \caption{\textbf{3DGSNav Real-World Demonstration.} The robot successfully navigates to locate the toilet. The reasoning process is simplified for clarity. The manipulator can move freely to support active perception.
    }
    \label{fig:real_toilet}
    \vspace{-4mm}
\end{figure}

\section{Introduction}
Object navigation (ObjectNav) is a core capability of embodied intelligence, requiring an agent to efficiently localize a target in unknown environments. This task requires an agent to jointly understand spatial structures and semantic information in complex scenes, as well as to perform long-horizon planning based on continuous observation. Recent works~\cite{trihelper,strive,zero} have incorporated vision–language models (VLMs) to support decision-making in zero-shot object navigation (ZSON), leveraging their strong contextual understanding and reasoning abilities. However, these approaches often fail to fully exploit the visual–spatial reasoning potential inherent in VLMs. Instead, they typically abstract environments into textual descriptions~\cite{navgpt,opennav} or construct semantic value maps~\cite{apexnav,vlfm} from current observations, thereby discretizing the continuous environment into semantic-level representations. While such formulations simplify the planning process, they weaken the spatial and structural relationships among objects. To mitigate this limitation, methods~\cite{sgnav,dragon} introduce 3D scene graphs to enhance relational modeling among objects. Nevertheless, the perception module remains a bottleneck for overall performance, preventing VLMs from fully leveraging their inherent spatial reasoning capabilities.

\emph{
"The ability of VLMs to generate travel guides, summarize lengthy documents, and edit images is truly impressive." -- VLM users
}

This observation highlights the strong planning, textual reasoning, and visual inference capabilities of VLMs. With the rapid advancement of VLMs, an important open problem is how to maximize their potential in ZSON planning while avoiding bottlenecks from low-level perceptual constraints. In this work, we propose 3DGSNav, a novel framework that constructs environmental representations using 3DGS through active perception, which are leveraged as persistent memory for VLMs. An opacity-driven viewpoint selection mechanism is introduced to enable the agent to actively acquire missing perceptual information, thereby improving overall sensing efficiency. Leveraging the efficient novel‑view synthesis (NVS) capability of 3DGS, historical observations are rendered from memory. A trajectory-guided free-viewpoint optimization strategy is then employed to generate informative first-person views (FPVs) that support the reasoning of the planner VLM~\cite{gemini3} used in our system. Furthermore, structured visual prompts are created from FPVs and a bird’s-eye view (BEV), with online annotations integrated into Chain-of-Thought (CoT) prompting to enhance the VLM spatial reasoning and long-horizon planning, enabling analysis and decision-making over candidate frontier points. To maintain lightweight and efficient navigation, a real-time object detector~\cite{yoloe} identifies potential targets, while the action-decision VLM~\cite{glm4.5v} performs target verification or selects appropriate actions when necessary. Selected actions are projected into the 3DGS memory to render novel viewpoints, enabling active viewpoint-based re-verification of potential targets and improving recognition reliability and accuracy. Our contributions include:
\begin{itemize}
\item We embed 3DGS as a memory representation for VLM, enabling efficient memory construction and retrieval via active perception and free-viewpoint optimization, and providing analyzable visual observations to support reasoning.

\item We design an annotation-empowered visual prompting framework that integrates structured visual prompts with CoT prompting to effectively elicit the spatial reasoning capabilities of VLMs for efficient ZSON, without relying on scene abstraction.

\item We conduct evaluations in real environments and on several datasets, and the experiments verify the advantages and efficacy of our approach.
\end{itemize}

\section{Related work}
\subsection{Object Navigation} 
Object-goal navigation is a fundamental task in embodied intelligence. Prior work primarily focuses on end-to-end methods, such as reinforcement learning~\cite{rl1,rl2,rl3} and imitation learning~\cite{ir1,habitatweb}, which learn direct mappings from sensory observations to actions in closed datasets or simulation. However, these approaches are often limited by sim-to-real gap and substantial computational costs, making real-world generalization challenging.

\subsection{Zero-Shot Object Navigation}

Benefiting from pretraining on large-scale datasets, large models have achieved remarkable progress in zero-shot understanding. To enable robots to recognize unseen target categories in unknown environments, prior work has explored incorporating large models into ObjectNav. Representative approaches~\cite{llmguide,l3mvn,openfmnav} construct semantic maps that textualize object information around frontier points and employ large language models (LLMs) to compute semantic similarity between candidate frontiers and the target object, thereby guiding exploration. Other methods~\cite{sgnav,dragon,hsg} build 3D scene graphs from semantic segmentation results to enrich scene representations and leverage LLMs to infer the next exploration location by reasoning over spatial and semantic relationships. Beyond LLM-based methods, VLFM~\cite{vlfm} integrates a VLM~\cite{blip} with online observations to continuously update a target-related similarity value map, improving exploration efficiency. ApexNav~\cite{apexnav} further enhances the robustness of semantic value map construction through rule-based optimization and the integration of additional auxiliary models. Recent advances in 3D scene representations have also opened new directions for embodied navigation. BeliefMapNav~\cite{beliefmapnav} adopts voxel-based maps with multi-level spatial semantics and LLM reasoning to estimate target confidence, while the work~\cite{3daware} leverages point clouds to build 3D semantic maps that capture fine-grained geometry and object relationships.

Despite these advances, most existing methods rely on semantic or textual scene abstractions that inevitably lose continuous spatial structure and fine‑grained geometric details. This forces high‑level decision making to depend primarily on semantic modules and symbolic representations, restricting effective navigation. To address this limitation, we replace scene abstractions with 3DGS-based representations to model the environment, enabling VLMs to fully exploit their inherent spatial understanding and reasoning capabilities for ZSON.

\subsection{LMMs for Spatial Reasoning}
Recent advancements in 3D large multimodal models (LMMs) have increasingly incorporated explicit geometric or reconstructive priors to enhance spatial reasoning capabilities, with applications spanning tasks such as 3D visual question answering (VQA) and 3D grounding~\cite{3dllm, ross3d}. Notably, 3D LLaVA~\cite{3dllava} and LL3DA~\cite{ll3da} focus on efficient modality bridging, utilizing superpoint pooling for compact 3D tokenization and query-based transformer aggregation for prompt-conditioned 3D understanding, respectively. To capture scene geometry and object relationships more effectively, Scene-LLM~\cite{scenellm} adopts a hybrid 3D visual feature representation with scene state updates, while another approach~\cite{inst3dlmm} explicitly models instance-level spatial relations. Recent video-centric methods further enhance spatiotemporal representations by integrating depth-derived or reconstruction-based 3D cues~\cite{video3dllm, vlm3r}. Evaluations on VSI-Bench~\cite{vsibench} indicate that recent video-language models~\cite{llavaonevision, improved, expanding} still face persistent challenges in visual spatial reasoning.

However, these approaches focus on short-horizon interaction, instruction following, or scene-level reasoning, where the model passively analyzes a given scene snapshot. ZSON necessitates long-horizon sequential decision-making under partial observability. In this dynamic setting, the agent must continuously integrate observations while preserving visual spatial consistency. To the best of our knowledge, we are the first to systematically incorporate 3DGS representations into VLMs to enable long-horizon, complex spatial reasoning.

\section{Approach}

\textbf{Task Definition.} In ZSON, given a natural language instruction $\mathcal{I}$ (e.g., “find a chair”), the agent is required to autonomously explore an unknown environment and navigate to a location within a distance $d_s$ of the target object and remaining there, all within a maximum of $\delta_{\max}$ steps. At each time step $t$, the agent receives an observation, which includes the RGB image $I^{t}$, depth map $D^{t}$, and pose $T_{cw}^t \in SE(3)$. The navigation policy then predicts an action $\mathcal{A}_t$. 

\begin{figure*}[t]
    \centering
    \includegraphics[scale=0.5]{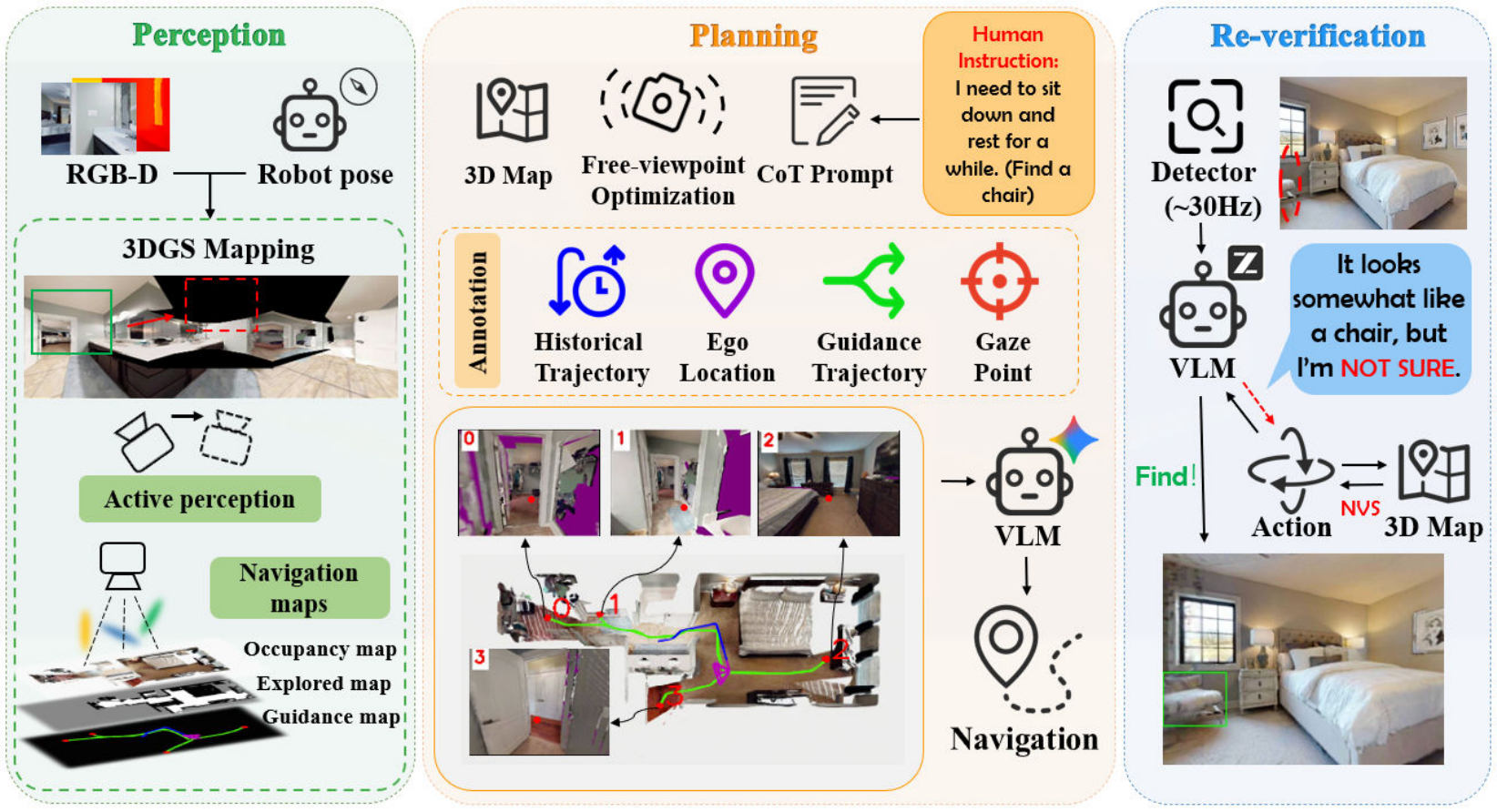}
    \caption{\textbf{System overview.} 
    The system builds navigation-oriented environment representations from robot poses and RGB-D observations via active perception. Free-viewpoint optimization and structured visual prompts guide VLM-based zero-shot navigation planning, while online object detection and viewpoint re-verification enable efficient target localization.
    }
    \label{fig:pipeline}
        \vspace{-6mm}
\end{figure*}

\textbf{Overview.} Figure~\ref{fig:pipeline} provides an overview of 3DGSNav. Given an agent pose and an RGB-D observation, environmental information is acquired via active perception, and navigation maps are constructed using 3DGS.
In the planning stage, a trajectory-guided free-viewpoint optimization method is introduced to render information-dense FPVs of candidate frontier points, which are augmented with explicit visual annotations. These structured visual prompts are then fed into the VLM, which performs spatial and semantic reasoning to select optimal exploration targets via CoT prompting.
During navigation, a real-time object detector is employed to efficiently detect potential targets. When detection results are ambiguous, the VLM is invoked to guide action selection by actively adjusting the rendering viewpoint, enabling the acquisition of more discriminative observations for target re-verification.

\subsection{Perception}
\subsubsection{Preliminaries} 
The 3D map is represented by a set of anisotropic Gaussian primitives $\textbf{G}$:
\begin{equation}
{\textbf{G}} = \{ {G_i}:(\mu_{i} ,o_{i},c_{i},\Sigma_{i} )|i = 1,...,N\},
\end{equation}
where each Gaussian primitive is defined by its position $\mu \in \mathbb{R}^{3}$, opacity $o \in [0,1]$, color $c \in \mathbb{R}^{3}$, and covariance matrix $\Sigma_i = RS{S^T}{R^T}$, where $S$ is a scale matrix and $R$ is a rotation matrix. 
Each Gaussian primitive $G_i$ is projected onto the image plane via the world-to-camera transformation $T_{cw}$ for rendering. The projected covariance matrix is computed as follows:
\begin{equation}
\Sigma '_{i} = JT_{cw}\Sigma_{i} T_{cw}^{T}{J^T},
\end{equation}
where $J$ denotes the Jacobian matrix of the affine approximation to the projective function. Therefore, the rendered color $\tilde I$, depth $\tilde D$, and opacity $\tilde O$ are computed as follows~\cite{3dgs}:
\begin{equation}
\{ \tilde I,\tilde D,\tilde O\}  = \sum\limits_{i = 1}^N {\{ {c_i},{z_i},{\mathrm{1}}\} {\sigma _i}} \prod\limits_{j = 1}^{i - 1} {(1 - {\sigma _j})},
\end{equation}
where the color of the i-th 3D Gaussian is given by $c_i$, $\sigma_{i}$ is the density term determined jointly by the Gaussian distribution and the learned opacity $o_{i}$, and $z_{i}$ denotes the corresponding depth value in the camera coordinate.

The 3D map is optimized in the backend under the following supervision:
\begin{align}
\label{equ:lgloss}
L_g &= \lambda_1 \big( \lambda \| \tilde I - I \|_2^2 + (1-\lambda)(1 - \textbf{SSIM}(\tilde I, I)) \big) \notag \\
    &\quad + \lambda_2 \| \tilde D - D \|_2^2,
\end{align}
where $\textbf{SSIM}$ is the structural similarity index measure~\cite{ssim}, and $\{\lambda_{*}\}$ are hyperparameters.

\subsubsection{Active perception} 
To improve perception efficiency, an active perception method is proposed to replace the inefficient passive rotational scanning. This approach leverages the flexibility of a virtual camera by rendering a panoramic opacity field to estimate observation completeness (details in Appendix~\ref{appendix:panoramic}). Subsequently, the DBSCAN algorithm $\Gamma$ is applied to cluster low-opacity regions, identifying representative discrete target viewpoints in the continuous viewpoint space. Since roll does not affect the observation process, the target viewpoint is computed from the region $\mathcal{S}$ with the largest low-opacity area and parameterized by pitch $\theta$ and yaw $\phi$. These angles are then used to guide the motion of the real camera, enabling it to actively cover visually missing regions.
\begin{equation}
(\theta^*, \phi^*) = \operatorname{Centroid}\left( \underset{\mathcal{C} \in \operatorname{\Gamma}(\{ (\theta, \phi), \tilde O(\theta, \phi) < \tau \})}{\arg\max} \mathcal{S}(\mathcal{C}) \right),
\end{equation}
where $\tau$ denotes the threshold used to measure low-opacity. This strategy not only avoids localization errors and redundant observations caused by mechanical rotation, but also significantly enhances exploration efficiency and robustness in long-term navigation.

\subsubsection{Frontiers} 
In the 3DGS space, we construct an exploration map by integrating opacity-based rendering and agent height information from a top-down perspective (details are provided in Appendix~\ref{appendix:exploremap}).
Frontier points $\mathcal{P}$, defined as the boundaries between explored and unexplored regions, are extracted from the explored map following~\cite{l3mvn}. To ensure that the extracted frontier points lie within observed regions, their positions are further refined based on Chebyshev distance.

Given the definition of frontiers, rendering FPVs from discrete frontier points often yields incomplete or local observations, whereas frontiers within the same region jointly capture the more environmental context. To effectively model the shared environmental information across multiple frontiers, we propose a spatial-structure adaptive frontier point clustering method. Specifically, a distance field is constructed from the geometric information encoded in the explored map, and local distance maxima are detected to identify core skeleton points. These skeleton points serve as adaptive seeds for a watershed segmentation algorithm, and the centroid of each segmented region is selected as a representative frontier point. This process enables unified modeling of redundant frontier information, producing higher-quality observations during rendering while reducing the computational overhead associated with analyzing isolated frontier points.

\subsubsection{Guidance trajectory} 
The guidance trajectory provides a principled reference for free‑viewpoint optimization and helps establish spatial awareness for VLM‑based planning. We design a cost function $C(n)$ based on the distance transform to guide the Dijkstra algorithm toward the shortest path to each frontier point, preventing the trajectory from closely adhering to obstacle boundaries. The total cost of node $n$ is defined as:
\begin{equation}
C(n) = C(n - 1) + \Delta L + \Omega (n),
\end{equation}
where $C(n-1)$ denotes the cost of the parent node, $\Delta L$ represents the step length that controls search efficiency, and $\Omega(n)$ is an exponential penalty term related to obstacle proximity. To encourage paths to stay away from high-risk regions, the penalty term is activated when the distance to the nearest obstacle $d(n)$ falls below a safety threshold $d_{s}$:
\begin{equation}
\Omega(n) = \omega \cdot \exp(2 \cdot (d_{s} - d(n))),
\end{equation}
where $ \omega$ is a weight. By constructing a virtual artificial potential field with a non-linear gradient, this function causes the cost to increase exponentially as the distance to obstacles decreases. The optimal guidance trajectory to the i-th frontier point $\mathcal{P}_i$ is obtained as $\mathcal{T}_i = \{n_1, n_2, \dots, n_j\}$, where each node $n_{i}$ satisfies the criterion of minimizing the accumulated cost $C(n)$.

\subsection{Planning}

\subsubsection{Virtual viewpoint initialization} 
Since the guidance trajectory $\mathcal{T}$ represents the shortest path, turns along the trajectory that are closest to a frontier point generally provide greater visibility of the frontier region. To focus on the frontier while capturing the surrounding environment, we design a point-based initialization strategy that leverages curvature $\hat{\kappa}_i$ and distance $\hat{d}_i$ to determine the initial virtual viewpoint position $\mathbf{p}^*$ by maximizing the weighted score:
\begin{equation}
\mathbf{p}^* = \arg \max_{\mathbf{p}_i \in {P}_{v}} \left[ \alpha \hat{\kappa}_i + (1 - \alpha) \hat{d}_i \right],
\end{equation}
where $\alpha$ is a weighting factor that balances the contributions of curvature and distance. The valid candidate set ${P}_v$ is constrained by a minimum safety distance $r_{min}$ and a maximum look-back distance $r_{max}$:
\begin{equation}
P_{v} = \left\{ \mathbf{p}_i \in \mathcal{T} \;\middle|\; r_{min} < \|\mathbf{p}_i - \mathcal{P}\| \leq r_{max} \right\}.
\end{equation}
This approach ensures that the initial position is neither too close, which could impair optimization efficiency, nor too distant, maintaining a manageable and stable solution space. Finally, the initial position $\mathbf{p}^*$ and the orientation toward the i-th frontier point $\mathcal{P}_i$ are combined to determine the initial pose $\xi^{*}_{i}$.

\subsubsection{Free-viewpoint Optimization}
The initial pose cannot handle all scenarios, with some viewpoints occluded and the VLM unable to reason about frontier information. To resolve this, camera pose estimation is formulated as a multi-constraint optimization problem to determine the optimal viewpoint. Given an initial pose $\xi^{*}_{k}$, the guidance trajectory $\mathcal{T}_{k}$, and the frontier point $\mathcal{P}_{k}$, the camera rotation quaternion $\mathbf{q}$ and translation $\mathbf{t}$ are iteratively optimized by minimizing a composite loss function:
\begin{equation}
\mathcal{L} = \lambda_{o}\mathcal{L}_{opa} + \lambda_{v}\mathcal{L}_{vis} + \lambda_{c}\mathcal{L}_{cos} + \lambda_{t}\mathcal{L}_{traj} .
\end{equation}

\textbf{Opacity loss.} The rendered FPV $\tilde I_{k}$ should balance observed and unobserved regions, as excessive unobserved areas can hinder analysis of the frontier environment. To quantify and regulate this balance, we introduce an opacity loss that measures the proportion of unobserved regions:
\begin{equation}
\mathcal{L}_{opa} = \mathbb{E}[1 - \tilde{O}],
\end{equation}
the $\mathbb{E}$ represents the averaging operation.

\textbf{Ray occlusion loss.} To ensure that the rendered FPV directly observes the frontier point, we model occlusion using a depth consistency constraint:
\begin{equation}
{{\cal L}_{vis}} = \mathbb{E}{[1 - Pr_{v}]}, Pr_{v} = \sigma ( {{{\tilde D}_{s}} - {z_{\mathcal{P}_{k}}}}) .
\end{equation}
Rays are cast from the camera center toward the frontier point and sampled at $N$ points along the ray. The occlusion probability is determined by a sigmoid $\sigma$ function, which compares the sampled depth $\tilde{D}_{s}$, obtained from the frontier point depth in the virtual camera, with the frontier point depth $z_{\mathcal{P}_{k}}$.

\textbf{View alignment loss.} To guarantee that the camera consistently gazes toward the frontier point $\mathcal{P}_{k}$, we compute the cosine similarity between the camera forward vector and the direction vector from the camera to the frontier. This loss term is defined as:
\begin{equation}
\mathcal{L}_{cos} = (1 - \cos^2(\theta)) \cdot \mathbb{E}[Pr_{v}],
\end{equation}
where $\theta$ is the angle between the two vectors. By introducing the occlusion probability $Pr_{v}$, the weight of this constraint is automatically reduced when the target is occluded.

\textbf{Guidance trajectory loss.} To enable the virtual camera to adjust its position near the guidance trajectory, we design a guidance trajectory loss. This loss calculates the minimum smoothed distance between the current camera position $\mathbf{t}$ and the trajectory point set $\mathbf{p}_{i} \in \mathcal{T}_k$:
\begin{equation}
\label{equ:trjloss}
\mathcal{L}_{traj} = -\frac{1}{\beta} \ln \sum_{i} \left( w \exp(-\beta {d_{i}}) \right), {d_{i}}=\|\mathbf{t} - \mathbf{p}_i\|_2,
\end{equation}
where $w$ is a weight factor based on the distance $d_{\mathcal{P}_{k}}$ from the camera to the frontier point, encouraging the camera to move closer during optimization. The weight is adaptively determined using the opacity loss:
\begin{equation}
w = {\exp( - \frac{{(1 - {\mathcal{L}_{opa}}){d_{\mathcal{P}_{k}}}}}{2})}.
\end{equation}
To prevent gradient backpropagation, the opacity loss $\mathcal{L}_{opa}$ is detached from $\mathcal{L}_{traj}$. The parameter $\beta$ serves as a smoothing factor, with further derivation details provided in the Appendix~\ref{appendix:beta}.


\subsubsection{Annotation}
To enhance the VLM's understanding of spatial structure and temporal evolution, online visual annotations are applied to both FPVs and the BEV, as illustrated in Figure~\ref{fig:real_toilet}. FPVs provide intuitive first-person visual observations, while the BEV supplies the VLM with global spatial context and historical information.

A gaze point is overlaid in the FPV to clearly indicate the region of interest of the current frontier point and to enhance the VLM spatial understanding, highlighting the navigation goal and its surrounding environment. Simultaneously, the BEV explicitly annotates the spatial layout of the frontier point along with the guidance trajectory of the agent, establishing a correspondence between the FPV and BEV and providing global spatial context.

To further support spatial memory and temporal reasoning, the BEV depicts the historical trajectory of the agent in blue and its current position as a purple triangle, providing continuous motion context. In addition, as the environment is gradually explored in ZSON tasks, unobserved regions are marked in purple in the FPV, guiding the VLM to differentiate between known and unknown areas and make informed exploration decisions. 

\subsubsection{Navigation Policy}
We integrate the FPV and BEV into a single composite image $\mathcal{X}$, which is combined with CoT $\mathcal{C}$ and fed into the VLM~\cite{gemini3} to determine the next frontier point: $\mathcal{P}^{*} = \mathcal{F}(\mathcal{I}, \mathcal{X}, \mathcal{C})$. The Fast Marching Method (FMM)~\cite{fmm} is then used to navigate the agent from its current position to the long-term goal.

\begin{figure}[b]
    \vspace{-6mm}
    \centering
    \includegraphics[scale=0.26]{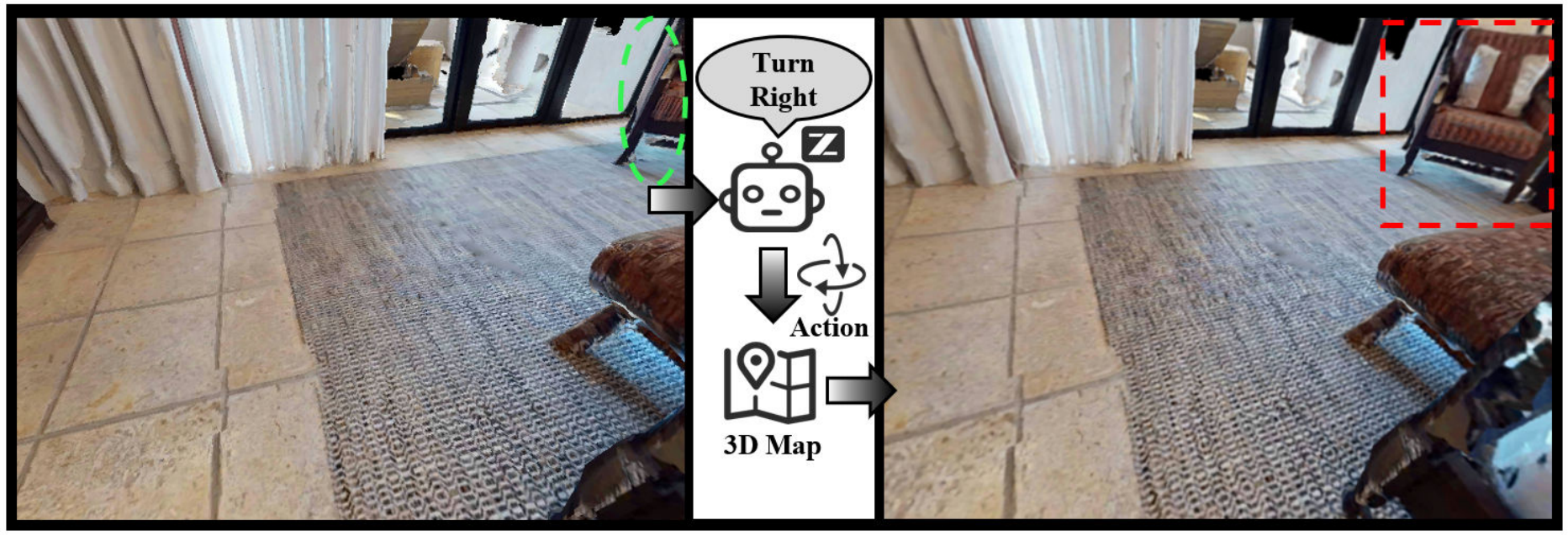}
    \caption{\textbf{Visualization of the action-decision VLM target re-verification.} The green circular bounding boxes indicate detections from the real-time detector, while the red bounding boxes denote the final detection results.
    }
    \label{fig:action}
\end{figure}

\subsection{Re-verification}
Detection accuracy is critical, as it provides the agent with the terminal signal for task completion. Methods~\cite{sgnav,cognav,strive} typically address uncertainty in detection accuracy by driving the agent to change its pose in order to acquire new observations. However, this strategy often incurs additional exploration costs and degrades overall navigation efficiency.

Therefore, as illustrated in Figure~\ref{fig:action}, we adopt a real-time detector~\cite{yoloe} that supports open-vocabulary detection via text prompts to perceive potential targets during navigation. The image containing detected potential targets is provided to the action-decision VLM~\cite{glm4.5v}, which reasons over the environmental structure and object occlusions to decide whether further actions should be taken from the action space
$\textbf{a}_t \in \{\textbf{forward (0.25 m)}, \allowbreak 
\textbf{backward(0.25 m)}, \allowbreak 
\textbf{turn left(10}^\circ\textbf{)}, \allowbreak 
\textbf{turn right(10}^\circ\textbf{)}, \allowbreak 
\textbf{leave}\}$. Instead of physically moving the robot, these selected actions are projected into the 3DGS space to render novel views for further target verification.

\begin{table*}[t]
    \centering
    \renewcommand{\arraystretch}{0.9}
    \caption{\textbf{Benchmark Comparisons.} Comparison of success rate of different methods on HM3Dv1, HM3Dv2, MP3D. Scene abstraction is defined as the representation of the environment in textual representations or semantic maps.}
    \label{tab:res}
    \setlength{\tabcolsep}{0.5mm}
    
    \begin{tabular}{cccccccccc}
        \hline
        \multirow{2}{*}{Method} & \multirow{2}{*}{Zero-shot} & \multicolumn{1}{c}{Scene} & \multirow{2}{*}{Unsupervised} & \multicolumn{2}{c}{HM3Dv1} & \multicolumn{2}{c}{HM3Dv2} & \multicolumn{2}{c}{MP3D} \\
               &           & abstraction &               & SR$\uparrow$ & SPL$\uparrow$ & SR$\uparrow$ & SPL$\uparrow$ & SR$\uparrow$ & SPL$\uparrow$ \\
        \hline
        Habitat-Web~\cite{habitatweb} & \XSolidBrush & \Checkmark  & \XSolidBrush & 41.5 & 16.0 & - & - & 31.6 & 8.5 \\
        SGM~\cite{sgm} & \XSolidBrush & \Checkmark & \XSolidBrush & 60.2 & 30.8 & - & - & 37.7 & 14.7 \\
        PSL~\cite{psl} & \Checkmark & \Checkmark & \XSolidBrush & 42.4 & 19.2 & - & - & 18.9 & 6.4 \\
        ZSON~\cite{zson} & \Checkmark & \XSolidBrush & \XSolidBrush & 25.5 & 12.6 & - & - & 15.3 & 4.8 \\
        ESC~\cite{esc} & \Checkmark & \Checkmark & \Checkmark & 39.2 & 22.3 & - & - & 28.7 & 14.2 \\
        OpenFMNav~\cite{openfmnav} & \Checkmark & \Checkmark & \Checkmark & 54.9 & 24.4 & - & - & 37.2 & 15.7 \\
        L3MVN~\cite{l3mvn} & \Checkmark & \Checkmark & \Checkmark & 50.4 & 23.1 & 36.3 & 15.7 & 34.9 & 14.5 \\
        VLFM~\cite{vlfm} & \Checkmark & \Checkmark & \Checkmark & 52.5 & 30.4 & 63.6 & 32.5 & 36.4 & 17.5 \\
        SG-NAV~\cite{sgnav} & \Checkmark & \Checkmark & \Checkmark & 54.0 & 24.9 & 49.6 & 25.5 & 40.2 & 16.0  \\
        BeliefMapsNav~\cite{beliefmapnav} & \Checkmark & \Checkmark & \Checkmark & 61.4 & 30.6 & - & - & 37.3 & 17.6 \\
        ApexNav~\cite{apexnav} & \Checkmark & \Checkmark & \Checkmark & 59.6 & 33.0 & \textbf{76.2} & 38.0 & 39.2 & 17.8 \\ \hline
        
        3DGSNav(Our) & \Checkmark & \XSolidBrush & \Checkmark & \textbf{80.0} & \textbf{51.79} & 75.0 & \textbf{44.19} & \textbf{43.63} & \textbf{21.31} \\
        \hline
    \end{tabular}
\end{table*}

\section{Experiments}


\begin{table}[t]
    \centering
    \setlength{\tabcolsep}{1mm}
    \caption{\textbf{Comparison of Different VLMs.} Random refers to a strategy where no VLM planning is performed, and frontier points are selected randomly.}
    \label{tab:qwen}
    \begin{tabular}{ccccccc}
        \hline
        \multirow{2}{*}{Method} &  \multicolumn{2}{c}{HM3Dv1} & \multicolumn{2}{c}{HM3Dv2} & \multicolumn{2}{c}{MP3D} \\
               &  SR$\uparrow$ & SPL$\uparrow$ & SR$\uparrow$ & SPL$\uparrow$ & SR$\uparrow$ & SPL$\uparrow$ \\
        \hline
        Random  & 50.0 & 16.85 & 64.81 & 32.67 & 25.0 & 4.91 \\
        Qwen32b  & 61.67 & 32.47 & 66.67 & 34.44 & 29.54 & 13.95 \\
        Qwen235b  & {65.0} & {37.6} & {68.51} & {37.15} & {31.81} & {14.61} \\
        3DGSNav(Our)  & \textbf{80.0} & \textbf{51.79} & \textbf{75.0} & \textbf{44.19} & \textbf{43.63} & \textbf{21.31} \\
        \hline
    \end{tabular}
\vspace{-6mm}
    
\end{table}

\begin{table}[t]
    \centering
    \setlength{\tabcolsep}{0.9mm}
    \caption{\textbf{Ablation study on viewpoints and annotations.} The virtual viewpoint is defined as a view directly oriented towards the frontier point from the agent current position, without viewpoints initialization or optimization.}
    \label{tab:alb}
    \begin{tabular}{ccccccc}
        \hline
        \multicolumn{1}{c}{Virtual} & 
        \multicolumn{1}{c}{Free-viewpoint} & 
        \multicolumn{1}{c}{Historical} & 
        \multicolumn{1}{c}{Gaze} & 
        \multicolumn{2}{c}{HM3Dv1}  \\ 
        viewpoints& optimization& trajectory& point & SR$\uparrow$ & SPL$\uparrow$  \\
        \hline
        \XSolidBrush & \Checkmark & \Checkmark & \Checkmark & 76.66 & 38.54  \\
        \Checkmark & \XSolidBrush & \Checkmark & \Checkmark & {78.33} & 47.7  \\
        \Checkmark & \Checkmark & \XSolidBrush & \Checkmark & 77.96 & 48.47  \\
        \Checkmark & \Checkmark & \Checkmark & \XSolidBrush & 78.0 & {48.59}  \\
        \Checkmark & \Checkmark & \Checkmark & \Checkmark & \textbf{80.0} & \textbf{51.79}  \\
        \hline
    \end{tabular}
    \vspace{-6mm}
\end{table}
\begin{table}[b]
    \centering
    \renewcommand{\arraystretch}{0.9}
    \vspace{-4mm}
    \setlength{\tabcolsep}{0.9mm}
    \caption{\textbf{Ablation study on re-verification, planner CoT, and frontier clustering.}}
    \label{tab:abl_2}
    \begin{tabular}{ccccccc}
        \hline
        \multicolumn{1}{c}{Re-verifi.} & \multicolumn{1}{c}{Re-verifi.}&
        \multicolumn{1}{c}{Planner}&
        \multicolumn{1}{c}{Frontier}& \multicolumn{2}{c}{HM3Dv1} &   \\
        model & action& CoT&clustering& SR$\uparrow$ & SPL$\uparrow$  \\
        \hline
        \XSolidBrush & \Checkmark& \Checkmark& \Checkmark & 53.75 & 33.19 \\
        \Checkmark & \XSolidBrush& \Checkmark& \Checkmark & 75.0  & {47.34}  \\
        \Checkmark & \Checkmark& \XSolidBrush& \Checkmark & {77.5} & 47.14 \\
        \Checkmark & \Checkmark& \Checkmark& \XSolidBrush & 62.5 & 37.68 \\
        \Checkmark & \Checkmark& \Checkmark& \Checkmark  & \textbf{80.0} & \textbf{51.79} \\
        \hline
    \end{tabular}
\end{table}
\begin{figure*}[t]
    \centering
    \includegraphics[scale=0.4]{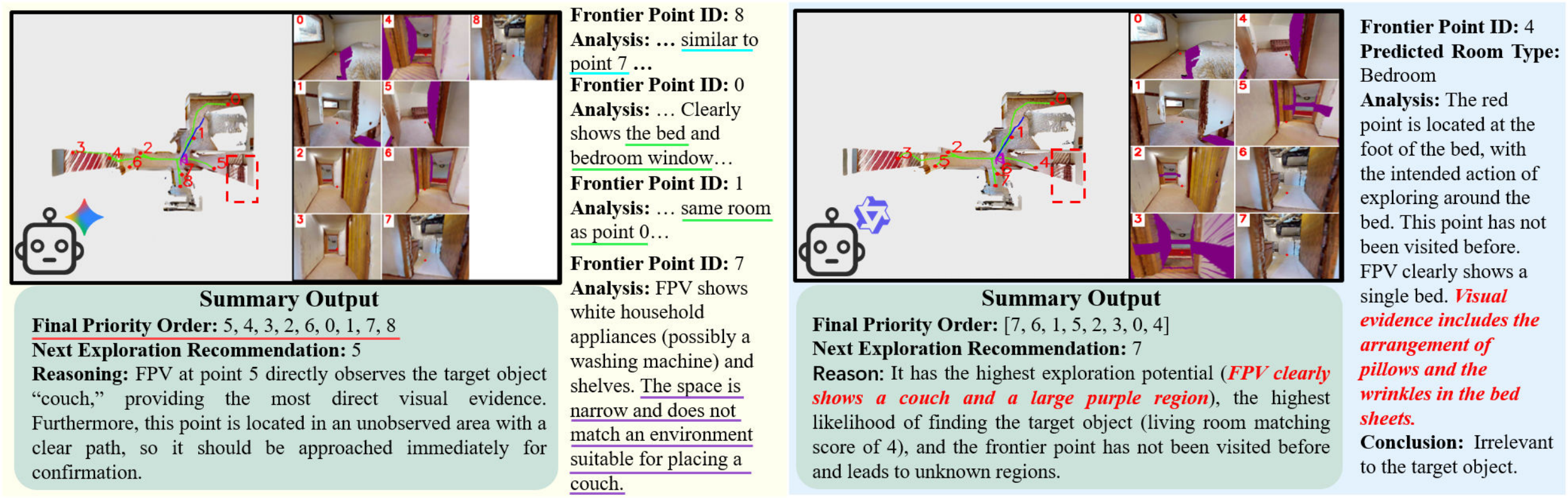}
    \caption{Comparison of self-explanations of Gemini3-Pro and Qwen3-235b-Thinking on ZSON tasks. The left and right images correspond to Gemini3-Pro and Qwen3-235b-Thinking, respectively. The target object is highlighted by a red bounding box.
    }
    \label{fig:vsqwen}
        \vspace{-4mm}
\end{figure*}

\subsection{Experiment Setup}
\textbf{Datasets.} We conduct experiments on three datasets in the Habitat~\cite{habitat} simulator. HM3Dv1~\cite{hm3dv1} contains 2000 episodes across 20 scenes with 6 goal categories. HM3Dv2~\cite{hm3dv2} includes 1000 episodes over 36 scenes with 6 goal categories. MP3D~\cite{mp3d} consists of 2195 episodes spanning 11 scenes with 21 goal categories.

\textbf{Metrics.} On all benchmarks, the $\delta_{\max}$ is set to 500 steps and $d_s$ is set to 1. We report Success Rate (SR(\%)) and Success weighted by inverse Path Length (SPL(\%)), which measures navigation efficiency relative to the optimal path length. The \textbf{best} results are highlighted in bold.

\textbf{Implementation Details.} In the Habitat setup, the robot can choose from a discrete action set $\mathcal{A}_t$: move forward (0.2 m), turn left (30°), turn right (30°), look up (40°), look down (40°), and stop. The agent’s camera is mounted at a height of 0.88 m. The navigation planning visual language model is Gemini3-Pro (Gemini3)~\cite{gemini3}. In the comparative experiments, we use Qwen3-VL-32b-Thinking (Qwen32b)~\cite{qwen3} and Qwen3-VL-235b-A22B-Thinking (Qwen235b)~\cite{qwen3}. For real-time object detection, we employ YOLOE~\cite{yoloe} with yoloe-11m-seg weights. The action-decision VLM is GLM-4.5v~\cite{glm4.5v}. All simulation experiments are conducted on a system equipped with an NVIDIA GeForce RTX 4060 Ti 16 GB GPU and an Intel i7-12700KF CPU. 
The 3DGS optimization uses $\{\lambda,\lambda_1,\lambda_2\}=\{0.2,1.0,0.7\}$. 
The active perception opacity threshold is set to $\tau = 0.3$. 
Guidance trajectory parameters are $\{\Delta L, d_s, \omega\}=\{4,10,5\}$. 
Virtual viewpoint initialization uses $\{\alpha,r_{min},r_{max}\}=\{0.7,1.0,4.0\}$. 
Free-viewpoint optimization adopts $\{\lambda_o,\lambda_v,\lambda_c, \lambda_t, \beta\}=\{0.01,1.0,0.01,0.1,5\}$ and runs for 40 iterations.


\subsection{Benchmark results}
Table~\ref{tab:res} reports the quantitative results of 3DGSNav on the HM3Dv1, HM3Dv2, and MP3D datasets with baselines. Compared with state-of-the-art methods, 3DGSNav achieves an average improvement of \textbf{13.5}\% in SR and \textbf{32.08}\% in SPL. Furthermore, compared with the RL-based ZSON~\cite{zson} method, which also does not rely on scene abstractions, 3DGSNav improves SR by an average of \textbf{203.01}\% and SPL by \textbf{320.11}\% on HM3Dv1 and MP3D, demonstrating the clear advantage of the VLM-based reasoning. In addition, relative to BeliefMapNav, which employs voxel representations and constructs 3D semantic maps, 3DGSNav achieves gains of \textbf{25.25}\% in SR and \textbf{51.65}\% in SPL. These results indicate that using 3DGS as a VLM accessible memory representation is more effective for spatial reasoning and navigation than scene abstractions.

\subsection{Ablation Study}
Due to space limitations, detailed analyses of VLM performance on ZSON tasks under our paradigm, as well as key findings and insights, are provided in the Appendix. Specifically, a bottleneck analysis of VLM performance on the ZSON task is presented in Appendix~\ref{appendix:failure}, the role of viewpoints and annotations in VLM reasoning is discussed in Appendix~\ref{appendix:annotation}, and the analysis of target re‑verification is included in Appendix~\ref{appendix:action}.
\subsubsection{Analysis of Different VLMs} 
To evaluate the effectiveness of VLMs in ZSON tasks, Table~\ref{tab:qwen} reports results from replacing the planning VLM~\cite{gemini3} with a random frontier selection strategy and two open-source VLMs. Gemini3 significantly outperforms the open-source Qwen235b model, achieving a \textbf{20.14}\% improvement in SR and a \textbf{31.25}\% improvement in SPL. 

To better understand this performance gap, self-explanations~\cite{vsibench} are employed to analyze the VLM reasoning process. In Figure~\ref{fig:vsqwen}, Gemini3 (left) exhibits advanced image and spatial understanding, as evidenced by the strong reasoning of \emph{"the space is narrow and does not match an environment suitable for placing a couch"}.
Furthermore, it can infer spatial relationships between different frontier points through the BEV, rather than performing isolated analysis or making inaccurate inferences. (e.g., \emph{Points 0 and 1 correspond to different FPV viewpoints but can be inferred to belong to the same room based on the BEV}.) Moreover, the final prioritization of the summary output further demonstrates that the current model possesses strong planning capabilities. 

\begin{figure}[t]
    \centering
    \includegraphics[scale=0.4]{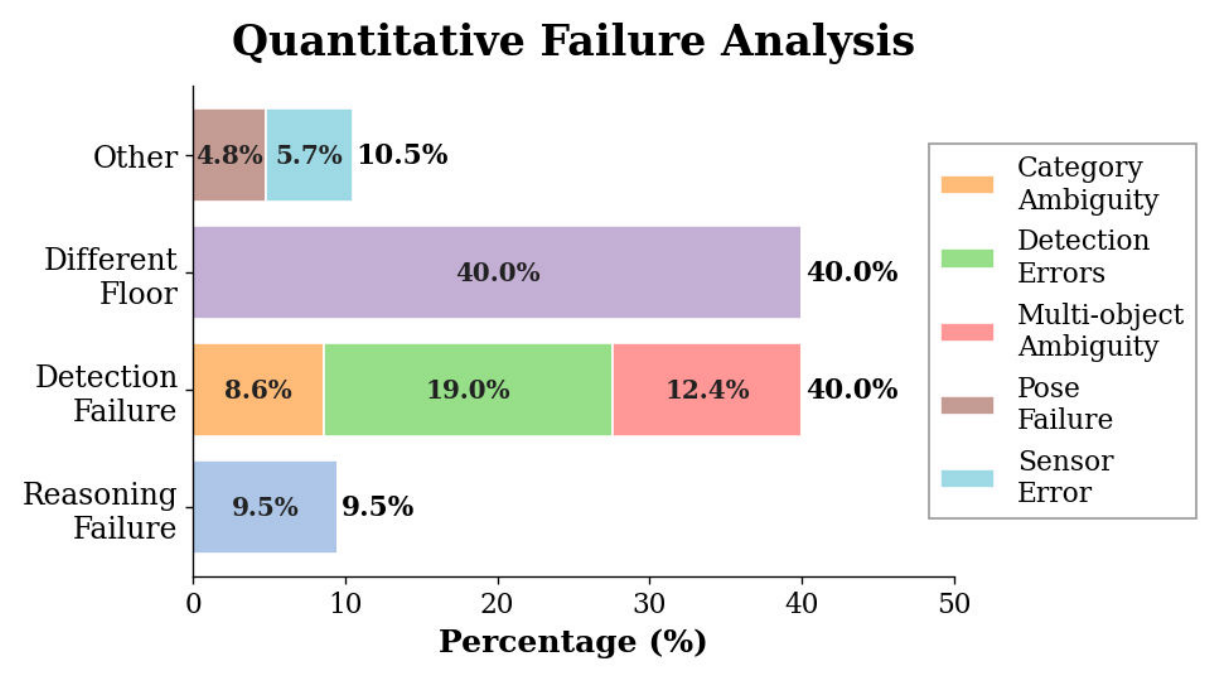}
     \vspace{-2mm}
    \caption{\textbf{Failure cause statistics.} Distribution of different failure types.
    }
    \label{fig:failure}
        \vspace{-6mm}
\end{figure}

Compared to random frontier point selection, 3DGSNav improves SR by \textbf{42.07}\% and SPL by \textbf{115.48}\%, while Qwen235b improves SR by \textbf{18.24}\% and SPL by \textbf{64.17}\%. However, its performance gains are smaller than those of Gemini3, mainly due to deficiencies in fundamental image understanding (e.g., \emph{"FPV clearly shows a couch and a large purple region."}) that lead to deviations in subsequent reasoning. Moreover, Figure~\ref{fig:failure} presents the distribution of failure types, showing that reasoning failures account for only 9.5\%, the lowest among all failure types. This finding further demonstrates the effectiveness of our paradigm in unlocking the potential of VLMs for ZSON tasks. For a more detailed analysis of the VLM, please refer to the Appendix~\ref{appendix:failure}.


\subsubsection{Viewpoints and annotations}
In Table~\ref{tab:alb}, we conduct an ablation study to evaluate the effectiveness of two key components: free viewpoints and visual annotations. Without virtual viewpoints, frontier perception is more likely to be occluded by walls or other obstacles, resulting in a \textbf{13.25}\% decrease in SPL. The annotated historical trajectory provides spatiotemporal guidance that reduces redundant backtracking, thereby improving navigation efficiency. More detailed analyses of viewpoints and annotations are provided in Appendix~\ref{appendix:annotation}.

\subsubsection{Re-verification}
In Table~\ref{tab:abl_2}, we conduct an ablation study to evaluate the effectiveness of the re-verification module, focusing on both the integration of the VLM and its action decision mechanism. Removing the re-verification VLM (i.e., relying solely on the real-time detector) results in a \textbf{32.81}\% drop in SR. Compared to the variant without VLM-driven action decisions, enabling action selection allows the VLM to perform more fine-grained reasoning over detection results, effectively exhibiting a CoT-like reasoning process. Moreover, by projecting the selected actions into the 3DGS representation to render novel viewpoints of the target, the VLM can mitigate misrecognition caused by partial or local observations, leading to more robust target verification. Additional visualizations are provided in Appendix~\ref{appendix:action}.

\subsubsection{Planner CoT and Frontier clustering}
In Table~\ref{tab:abl_2}, the results show that CoT is essential for effective reasoning in ZSON tasks. Moreover, frontier point clustering aggregates dispersed frontiers to enrich free‑viewpoint rendering, yielding to a \textbf{21.87}\% improvement in SR. Details of the CoT prompts are provided in Appendix~\ref{appendix:cot}.

\subsection{Real‑robot Experiments}
Our method achieves a \textbf{69.44}\% SR on a quadruped robot in real‑world office and hotel environments. More details and results are provided in Appendix~\ref{appendix:realworld}.




\section{Conclusion}
In this work, we presented 3DGSNav, a novel framework for ZSON that leverages 3DGS as a memory representation to enhance VLM reasoning. By combining active perception, trajectory-guided free-viewpoint rendering, and structured visual prompts with CoT prompting, our method enables efficient spatial reasoning and planning. 3DGSNav achieves strong performance on both real‑world robotic platforms and public benchmarks. Future work will focus on improving generality to support a broader range of tasks.



\nocite{langley00}

\bibliography{example_paper}
\bibliographystyle{icml2026}

\newpage
\appendix
\onecolumn
\section{Appendix Outline}
In these supplementary materials, we provide:
\begin{itemize}
\item Panoramic Rendering~\ref{appendix:panoramic};
\item Exploration Map Generation~\ref{appendix:exploremap};
\item Role of $\beta$ in Guidance Trajectory Loss~\ref{appendix:beta};
\item Analysis of VLM Failure Cases~\ref{appendix:failure};
\item Analysis of Viewpoints and Visual Annotations~\ref{appendix:annotation};
\item Analysis of Re‑verification Cases~\ref{appendix:action};
\item Runtime Analysis~\ref{appendix:runtime};
\item Real‑World Experimental Details~\ref{appendix:realworld};
\item Chain‑of‑Thought Prompt Design~\ref{appendix:cot};

\end{itemize}

\section{Panoramic Rendering}
\label{appendix:panoramic}
We exploit the flexibility of the virtual camera, which allows arbitrary adjustment of intrinsic parameters, by setting the horizontal field of view (hFoV) to 120° and the vertical field of view (vFoV) to 150°. To achieve full panoramic coverage, we render $2\pi/\text{hFoV}$ evenly spaced views centered around the agent. Each rendered image has a resolution of 120 × 150 pixels. The corresponding camera intrinsic matrix is computed as follows:
\begin{equation}
    K = \begin{bmatrix} f_x & 0 & c_x \\ 0 & f_y & c_y \\ 0 & 0 & 1 \end{bmatrix},f_x=f_y=\frac{W}{2 \tan(\frac{\theta_h}{2})},c_x=c_y= \frac{W}{2}
\end{equation}
The panoramic image is obtained by horizontally concatenating all rendered views, as shown in Figure~\ref{fig:pipeline}. The same procedure is applied to generate panoramic depth fields and opacity fields.

\section{Exploration Map Generation}
\label{appendix:exploremap}
We remove Gaussian primitives corresponding to the ceiling and the floor based on the height of the agent. A top down rendering is then performed to obtain the global opacity map of the scene. Regions with opacity values below a threshold $\tau$ are regarded as obstacles. Meanwhile, the floor is identified and treated as a traversable area. Finally, the obstacle map and the traversable map are combined to construct the exploration map.

\section{Role of $\beta$ in Guidance Trajectory Loss}
\label{appendix:beta}
To better understand the role of $\beta$ in the optimization process, we derive the formulation:
\begin{equation}
\mathcal{L}_{traj} = -\frac{1}{\beta} \ln \sum_{i} \left( w e^{-\beta {d_{i}}} \right), {d_{i}}=\|\mathbf{t} - \mathbf{p}_i\|_2.
\end{equation}

For simplicity, we set $w=1$, and the loss becomes:
\begin{equation}
\mathcal{L}_{traj} = -\frac{1}{\beta} \ln \left( \sum_{i} e^{-\beta d_i} \right).
\end{equation}
We compute the gradient of the loss function with respect to the camera position $\mathbf{t}$:
\begin{equation}
\frac{\partial \mathcal{L}_{traj}}{\partial \mathbf{t}} = -\frac{1}{\beta} \cdot \frac{1}{\sum_{j} e^{-\beta d_j}} \cdot \sum_{i} \left( e^{-\beta d_i} \cdot (-\beta) \frac{\partial d_i}{\partial \mathbf{t}} \right).
\end{equation}
Since $\frac{\partial d_i}{\partial \mathbf{t}} = \frac{\mathbf{t} - \mathbf{p}_i}{d_i}$, which corresponds to the unit vector pointing from point $\mathbf{p}_i$ to the camera position $\mathbf{t}$, we can cancel the factor $\beta$ to obtain:
\begin{equation}
\frac{\partial \mathcal{L}_{traj}}{\partial \mathbf{t}} = \sum_{i} \left( \frac{e^{-\beta d_i}}{\sum_{j} e^{-\beta d_j}} \right) \frac{\mathbf{t} - \mathbf{p}_i}{d_i}.
\end{equation}
By defining $\alpha_i = \frac{e^{-\beta d_i}}{\sum_{j} e^{-\beta d_j}}$, the gradient can be written as:
\begin{equation}
\frac{\partial \mathcal{L}_{traj}}{\partial \mathbf{t}} = \sum_{i} \alpha_i \cdot \mathbf{u}_i,
\end{equation}
where $\mathbf{u}_i$ denotes the unit direction vector.

The coefficient $\alpha_i$ follows a softmax distribution, where $\beta$ controls the concentration of the distribution. The hyperparameter $\beta$ serves as a smoothing factor that modulates the sparsity of the influence of trajectory points. By adjusting $\beta$, we can transition the optimization from a global centroid attraction ($\beta \to 0$) to a local point-wise alignment ($\beta \to \infty$). Therefore, a moderate $\beta$ value ensures a smooth optimization landscape, allowing the camera pose to converge robustly toward the manifold defined by the reference trajectory $\mathcal{T}$.

\section{Analysis of VLM Failure Cases}
\label{appendix:failure}
\begin{figure}[t]
    \centering
    \includegraphics[scale=0.4]{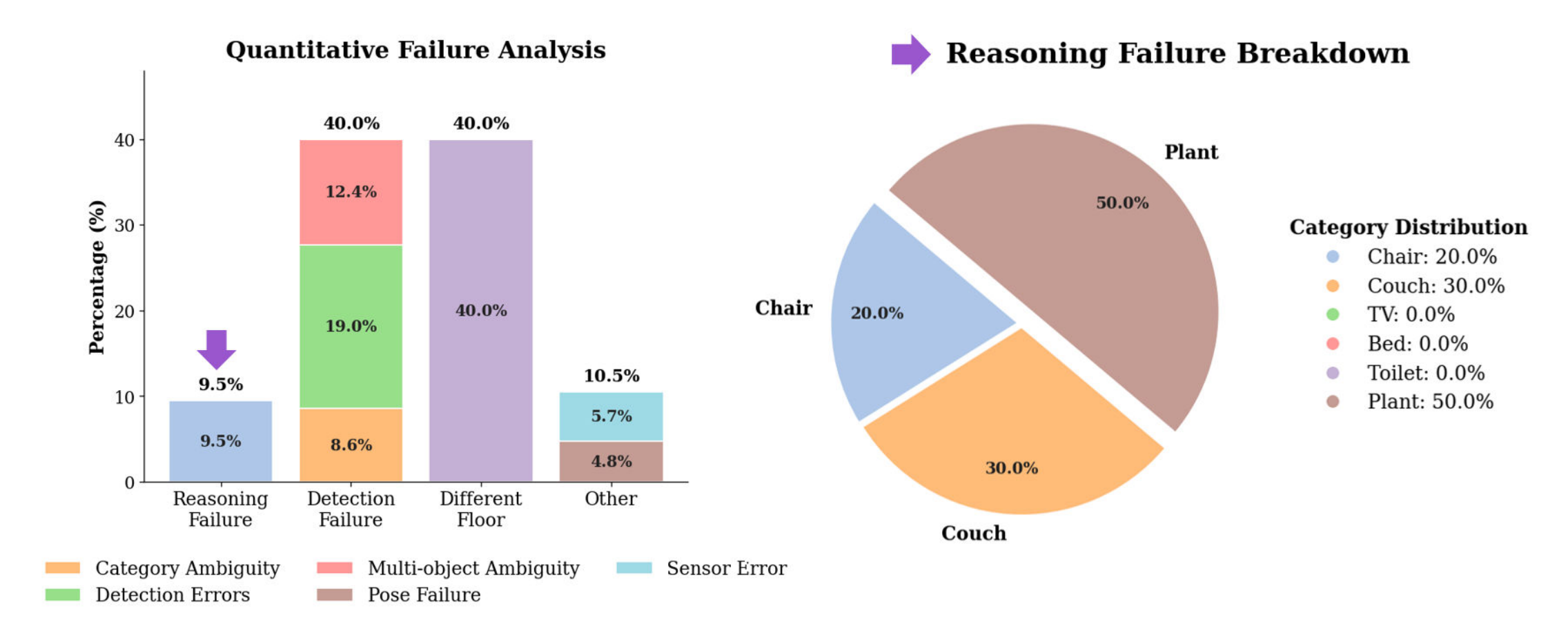}
    \caption{\textbf{Failure Case Statistics and Reasoning Failure Breakdown.} Failure cases include reasoning failure, detection failure (e.g., ambiguous categories like gaming chair vs. chair, multiple targets like multiple chairs, and detection errors such as false positives/negatives), different floor, and other issues such as sensor failure (invalid image/depth) and pose failure. The six categories in the reasoning failure breakdown correspond to the target object classes in HM3Dv1 and HM3Dv2. The ratios denote the failure rate specifically for episodes involving each target object.
    }
    \label{fig:failure1}
\end{figure}
A statistical analysis of different failure types is presented in Figure~\ref{fig:failure1}. The results indicate that the primary cause of task failure lies in the final target detection stage, which can be further divided into three categories. Among these categories, category ambiguity and multi-target ambiguity originate from the simulator’s decision mechanism.

Specifically, category ambiguity occurs when the issued instruction refers to an object category such as “chair.” Although the environment contains valid subcategories, including wooden chairs and gaming chairs, and the agent correctly recognizes them as semantically consistent with the instruction (as shown in Figure~\ref{fig:redpoint}, middle), the simulator still labels the episode as a failure.

In addition, multi-target ambiguity refers to scenarios involving multiple instances of the same object category (e.g., chairs near a bar counter and chairs in a kitchen). In such cases, the simulator considers only the kitchen chair as the correct match, while treating the chair near the bar counter as an incorrect target. Because the instruction provided to the simulator specifies only the category “chair” without additional qualifiers, the task fails due to the inability to disambiguate among multiple valid targets.
Furthermore, detection errors account for 19\% of failures, primarily arising from constrained local perception and degraded novel‑view quality after action execution. This result further indicates that high‑quality and robust novel‑view synthesis is not only critical for planning and decision-making, but also essential for ensuring accurate and reliable target detection.

Notably, reasoning failures constitute the smallest proportion of overall failures, suggesting that our proposed paradigm successfully unlocks the potential of the VLM for ZSON. Finally, because the navigation policy is based on two-dimensional information, 3DGSNav is currently unable to perform cross-floor navigation, which remains an important direction for future improvement.

\begin{figure}[t]
    \centering
    \includegraphics[scale=0.52]{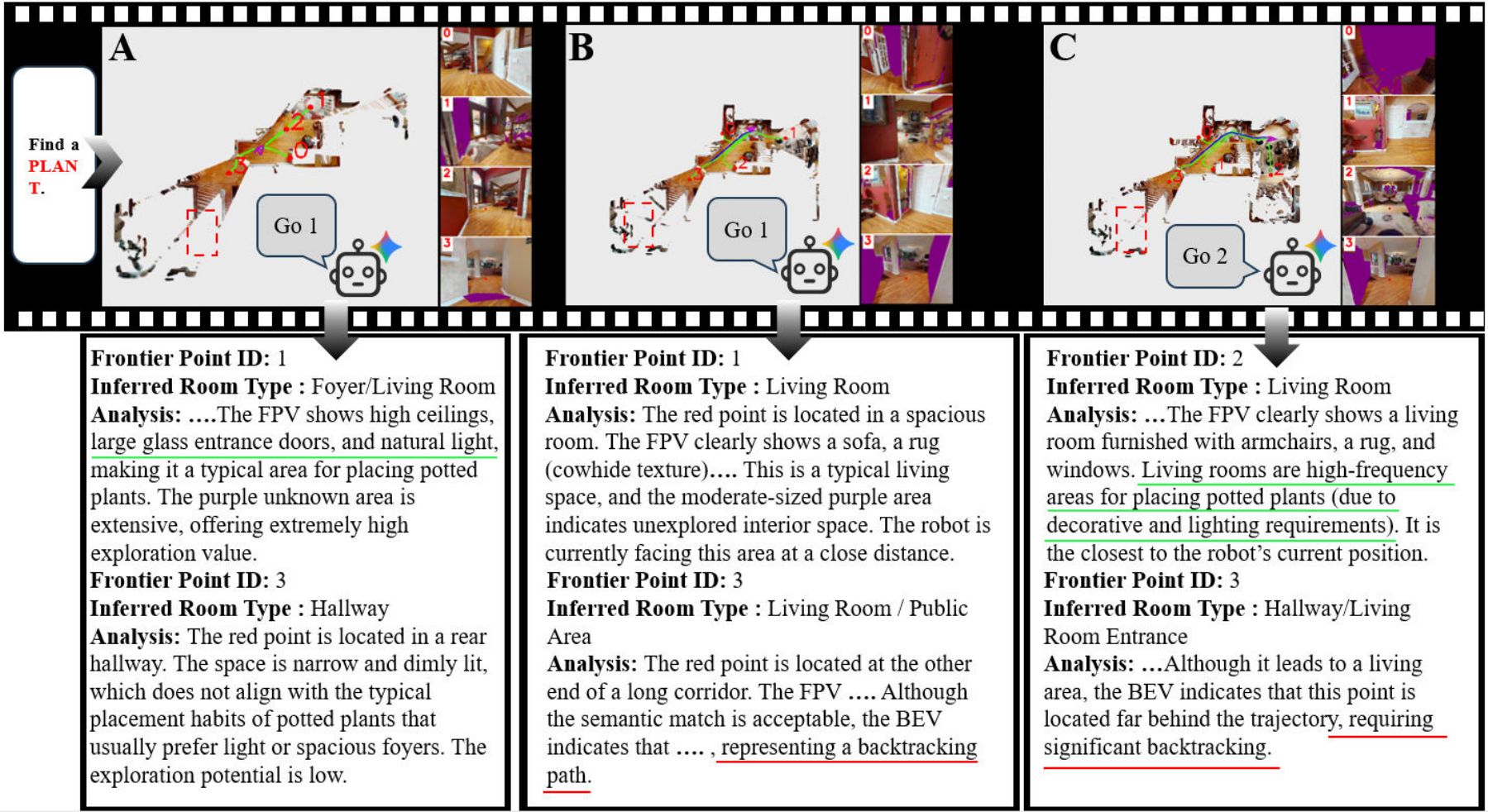}
    \caption{\textbf{VLM reasoning failure example 1.} The target object is a plant and is highlighted with a red bounding box.
    }
    \label{fig:failure_plant}
\end{figure}
\subsection{Analysis of VLM Reasoning Failures}
To quantify and identify the bottlenecks of the proposed paradigm in ZSON tasks, we conduct an in-depth analysis of several failure cases. 

\subsubsection{Weak Target Priors and the Backtracking Issue}
When target objects lack distinctive spatial priors (e.g., plants appearing in arbitrary locations, unlike beds confined to bedrooms), the agent must rely on weak cues such as lighting and layout. Although this strategy is reasonable under ambiguity (Fig.~\ref{fig:failure_plant} A), historical observations (blue trajectory in Fig.~\ref{fig:failure_plant}) often dominate decision-making.

When multiple cues provide comparable evidence, backtracking becomes penalized by the reward mechanism. As a result, the agent tends to avoid revisiting explored areas and continues forward, increasing the likelihood of failure. This trend is corroborated by the reasoning failure breakdown in Fig.~\ref{fig:failure1}. Category-wise analysis further reveals that most failures occur on weak-prior targets, such as plants, whereas objects with strong spatial and functional priors (e.g., TVs, beds, and toilets) exhibit near-zero failures.

\begin{figure}[t]
    \centering
    \includegraphics[scale=0.52]{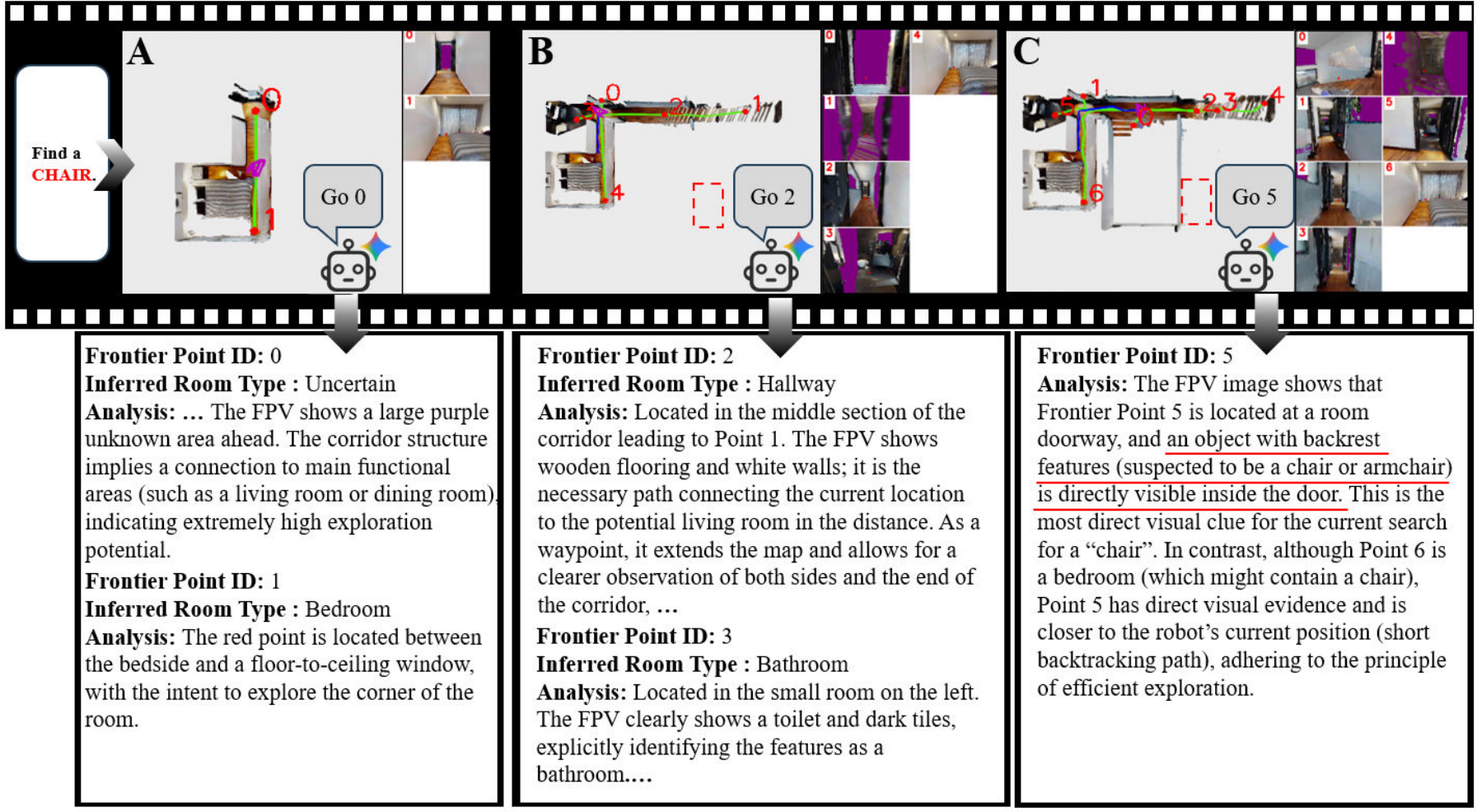}
    \caption{\textbf{VLM reasoning failure example 2.} The target object is a chair and is highlighted with a red bounding box.
    }
    \label{fig:failure_chair}
\end{figure}

\subsubsection{Visual perception Issue}
In Figure~\ref{fig:failure_chair} C, a hallucination occurs at Frontier~5, which corresponds to Frontier~3 in Figure~\ref{fig:failure_chair} B, where the agent infers a potential presence of a chair.

\subsubsection{exploration Issue}
\begin{figure}[b]
    \centering
    \includegraphics[scale=0.35]{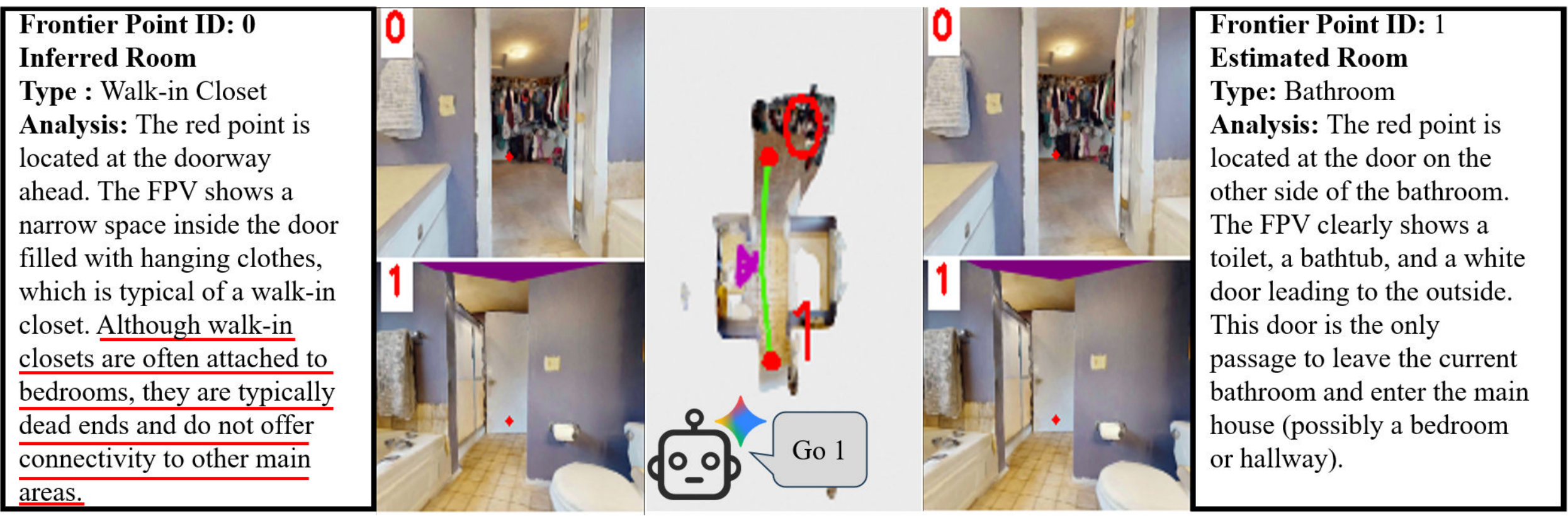}
    \caption{\textbf{VLM reasoning failure example 3.} The target object is a bed.
    }
    \label{fig:failure_bed}
\end{figure}

\begin{figure}[t]
    \centering
    \includegraphics[scale=0.42]{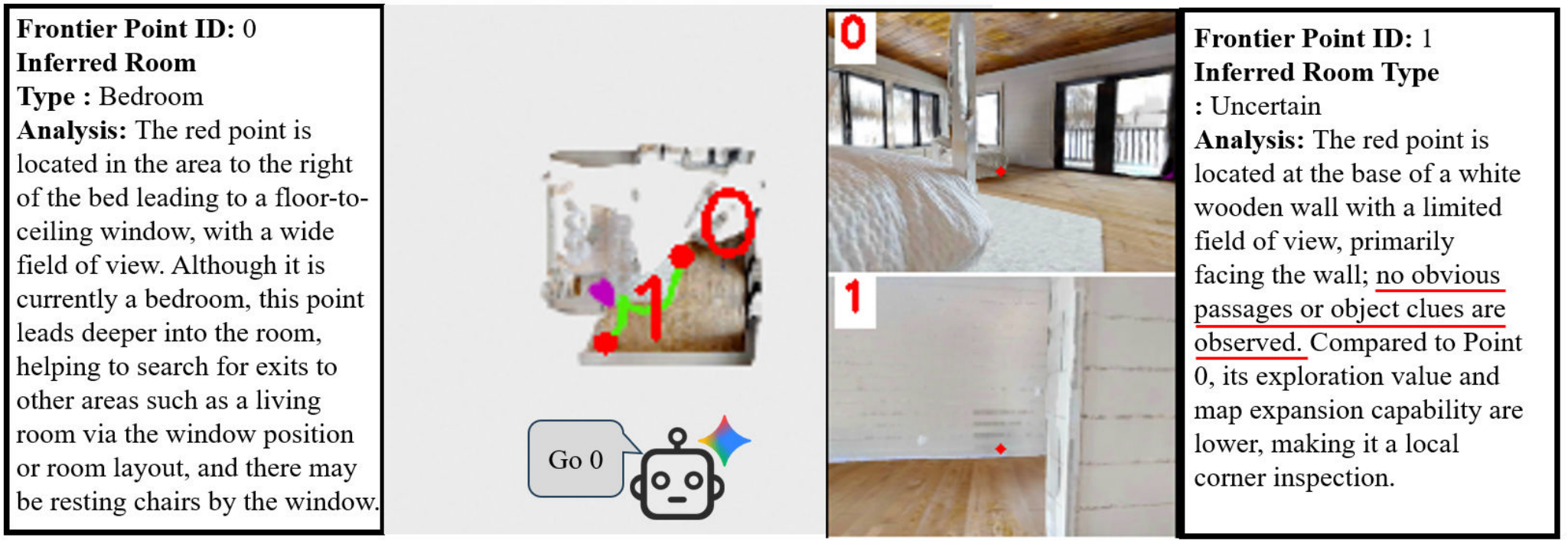}
    \caption{\textbf{VLM reasoning failure example 4.} The target object is a couch.
    }
    \label{fig:failure_sofa}
\end{figure}

Active exploration is a critical capability that agents must possess in ZSON tasks. However, extensive experimental evidence suggests that VLMs tend to exhibit a conservative bias when exploring unknown environments, preferring regions with even weak visual or semantic cues over proactively investigating entirely unexplored areas. The hallucination observed in Figure~\ref{fig:failure_chair} may stem from insufficient environmental information provided by other frontier points, which can cause the VLM to produce seemingly plausible yet ultimately incorrect justifications when selecting Frontier~5. Furthermore, Figures~\ref{fig:failure_bed} and~\ref{fig:failure_sofa} further highlight the challenges VLMs face in leveraging BEV information to assess whether frontier points correspond to previously unexplored spaces.


\begin{table*}[t]
    \centering
    \renewcommand{\arraystretch}{0.9}
    \caption{\textbf{Ablation study on viewpoints and annotations.} The virtual viewpoint is defined as a view directly oriented towards the frontier point from the agent current position, without viewpoints initialization or optimization.}
    \label{tab:alb_1}
    \begin{tabular}{cccccccccc}
        \hline
        \multicolumn{1}{c}{Virtual} & 
        \multicolumn{1}{c}{Free-viewpoint} & 
        \multicolumn{1}{c}{Historical} & 
        \multicolumn{1}{c}{Gaze} & 
        \multicolumn{2}{c}{HM3Dv1} & 
        \multicolumn{2}{c}{HM3Dv2} & 
        \multicolumn{2}{c}{MP3D} \\ 
        viewpoints& optimization& trajectory& point & SR & SPL & SR & SPL & SR & SPL \\
        \hline
        \XSolidBrush & \Checkmark & \Checkmark & \Checkmark & 76.66 & 38.54 & 72.22 & 38.26 & 30.3 & 16.68 \\
        \Checkmark & \XSolidBrush & \Checkmark & \Checkmark & {78.33} & 47.7 & {74.07} & 36.72 & 39.39 & 18.32 \\
        \Checkmark & \Checkmark & \XSolidBrush & \Checkmark & 77.96 & 48.47 & 72.22 & 37.51 & 36.36 & 17.32 \\
        \Checkmark & \Checkmark & \Checkmark & \XSolidBrush & 78.0 & {48.59} & 72.22 & {38.52} & {42.42} & {20.06} \\
        \Checkmark & \Checkmark & \Checkmark & \Checkmark & \textbf{80.0} & \textbf{51.79} & \textbf{75.0} & \textbf{44.19} & \textbf{43.63} & \textbf{21.31} \\
        \hline
    \end{tabular}
    \vspace{-4mm}
\end{table*}
\begin{figure*}[t]
    \centering
    \includegraphics[scale=0.5]{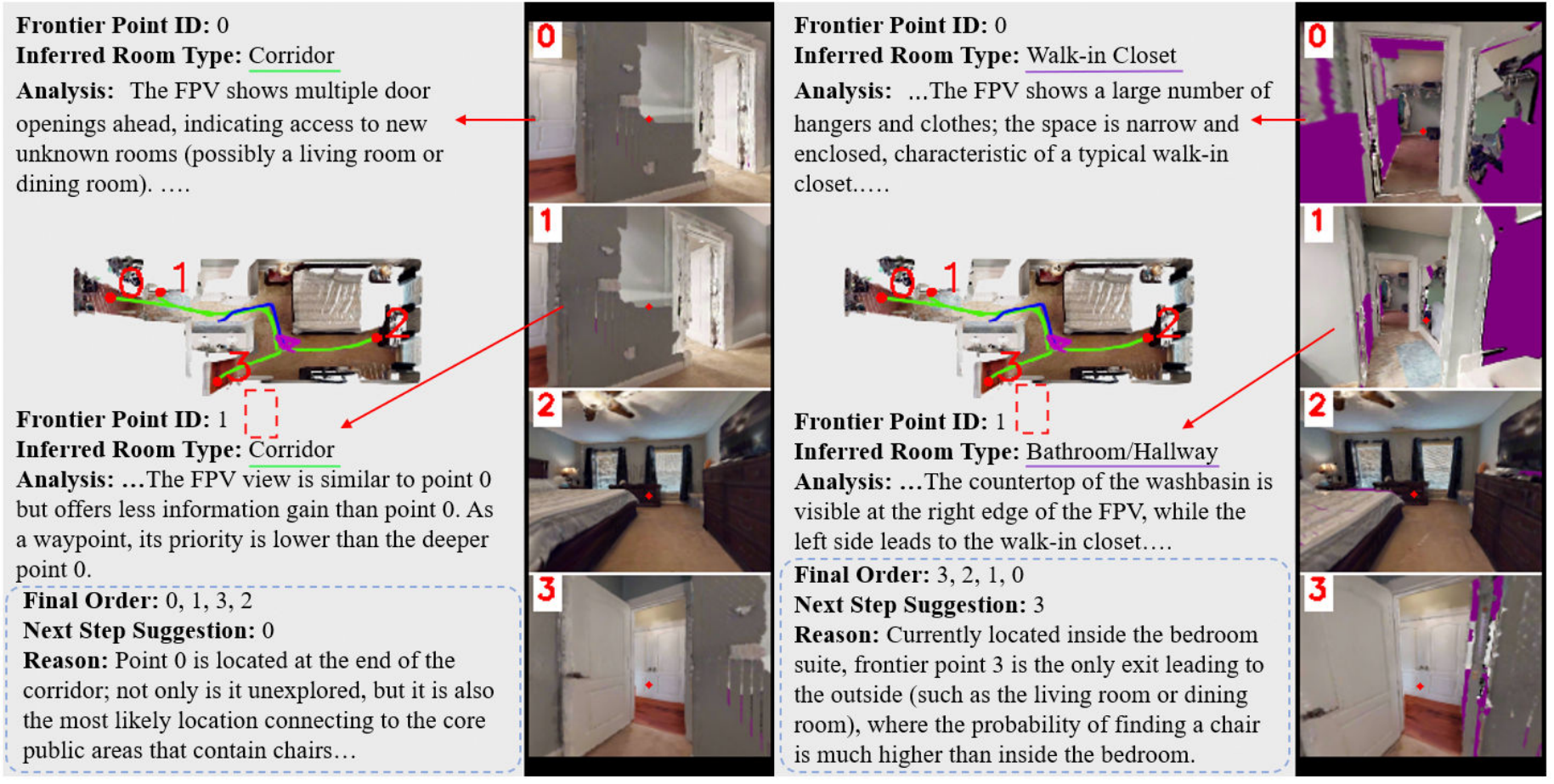}
    \caption{\textbf{Ablation Study on Virtual Viewpoints.} The left image shows rendering without virtual viewpoints, while the right image shows rendering based on optimized virtual viewpoints. The target object is a chair and is highlighted with a red bounding box.}
    \label{fig:vsview}
\end{figure*}
\begin{figure*}[t]
    \centering
    \includegraphics[scale=0.4]{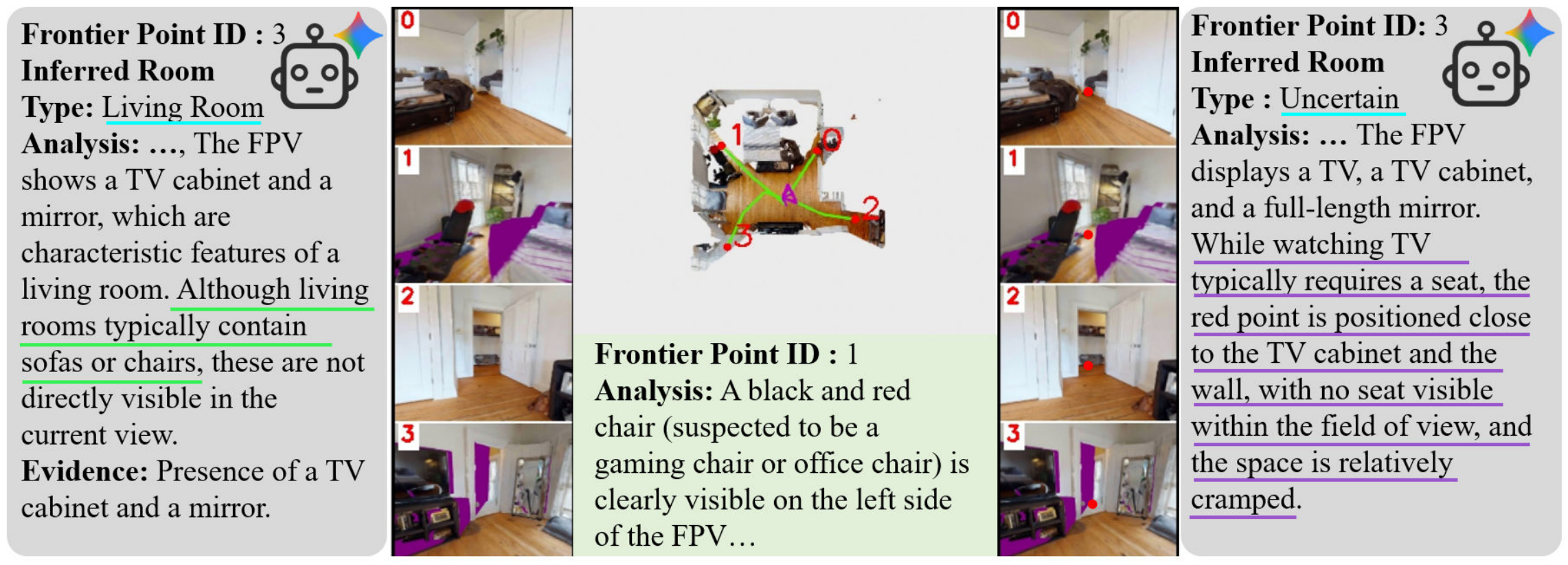}
    \caption{\textbf{Presence of gaze point visualization example 1.} The red point is the gaze point.}
    \label{fig:redpoint}
\end{figure*}
\begin{figure*}[t]
    \centering
    \includegraphics[scale=0.5]{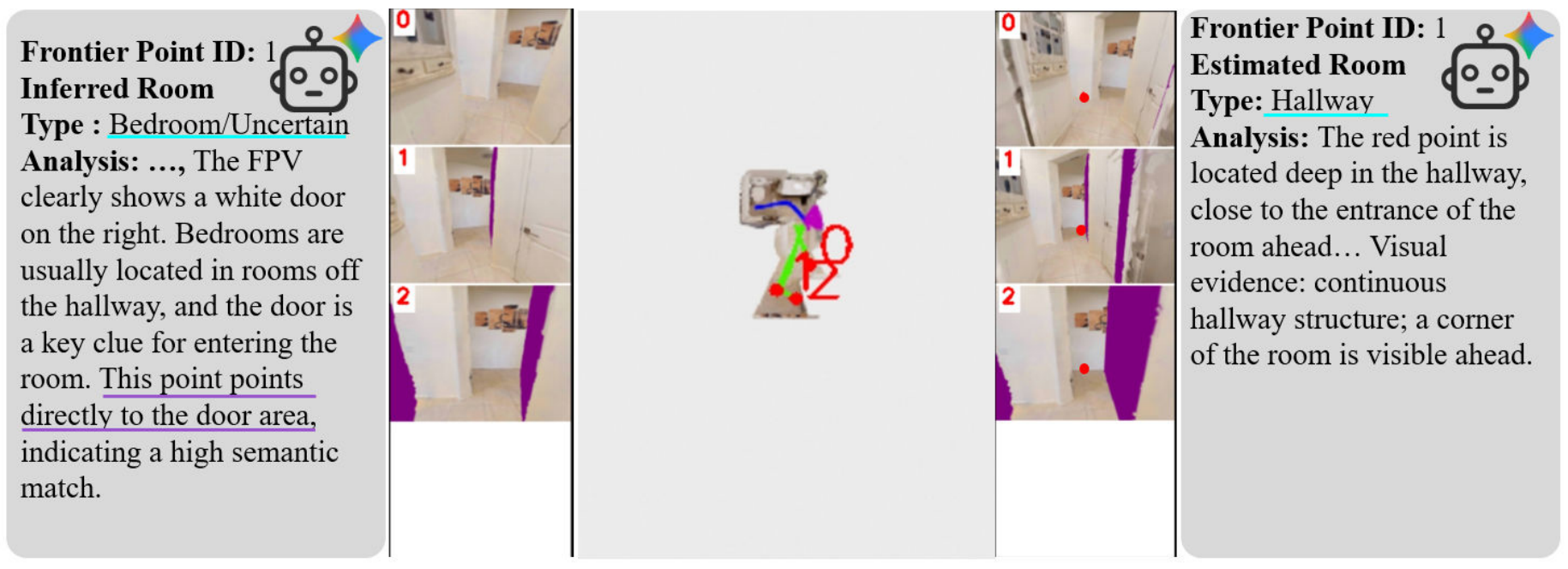}
    \caption{\textbf{Presence of gaze point visualization example 2.} The red point is the gaze point.}
    \label{fig:redpoint1}
\end{figure*}

\section{Analysis of Viewpoints and Visual Annotations}
\label{appendix:annotation}
\subsection{Viewpoint Analysis}
Figure~\ref{fig:vsview} presents an ablation study examining the impact of virtual viewpoints. For Frontiers 1 and 2, frontier information is entirely occluded by walls when virtual viewpoints are not employed. In contrast, virtual viewpoint rendering produces FPVs that successfully reveal relevant frontier cues. With access to these additional visual information, the VLM can more accurately infer the room type associated with each frontier point.

Furthermore, Table~\ref{tab:alb_1} indicates that removing virtual viewpoints results in the largest performance degradation in SPL among all evaluated components, with a reduction of up to \textbf{25.47}\%. This substantial decline in navigation efficiency causes many episodes to exceed the maximum step limit, ultimately leading to a \textbf{10.59}\% drop in SR.

Free‑viewpoint optimization helps overcome constraints imposed by limited initial pose flexibility, thereby increasing the likelihood of successfully observing informative frontier regions. Overall, these findings demonstrate that renderings generated via 3DGS can be effectively interpreted and exploited by VLMs to support more accurate reasoning and efficient navigation.

\subsection{Annotation Visualization Analysis}
\begin{figure*}[b]
    \centering
    \includegraphics[scale=0.5]{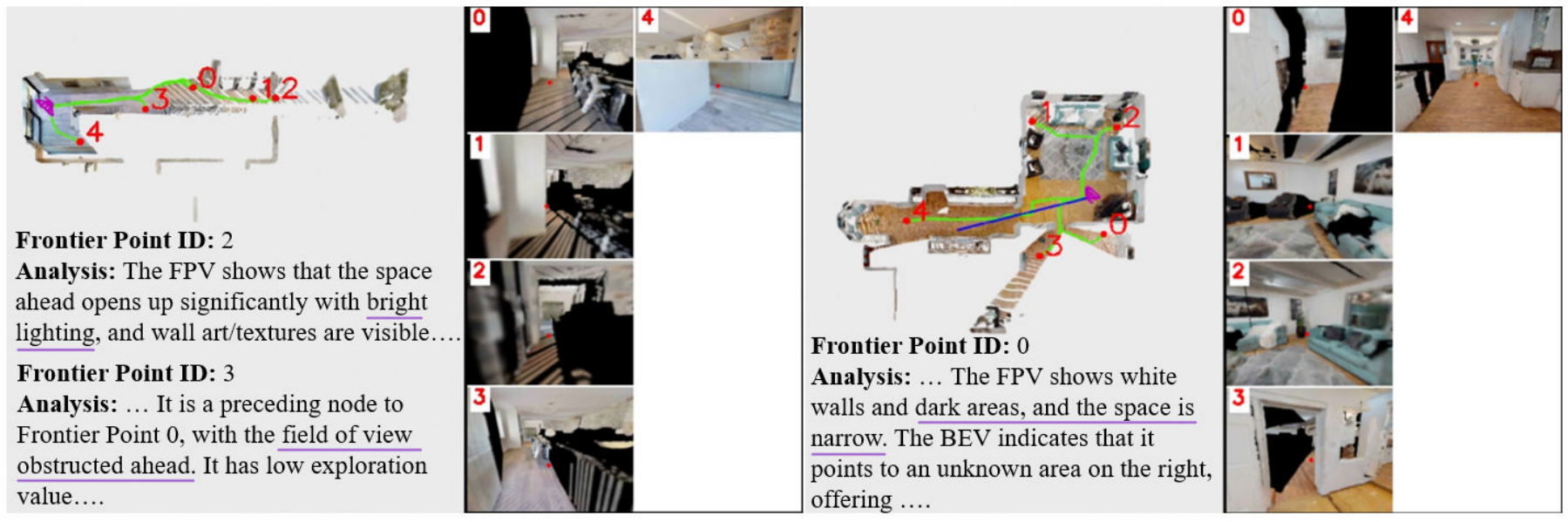}
    \caption{\textbf{FPV unobserved representation.}}
    \label{fig:black}
\end{figure*}

\subsubsection{Gaze Point}
As shown in Figure~\ref{fig:redpoint} (left), without gaze-point guidance, the VLM processes the entire image holistically. Distracted by irrelevant objects such as the television and mirror, it incorrectly classifies Frontier~3 as part of a living room and subsequently bases its object-likelihood reasoning on this erroneous room type. In contrast, the result in Figure~\ref{fig:redpoint} (right) labels the room type as uncertain, preventing incorrect room-based inference. This comparison demonstrates that gaze points effectively steer the VLM’s attention toward the spatial region surrounding the frontier, thereby improving its spatial understanding.

Similarly, in Figure~\ref{fig:redpoint1} (left), the absence of gaze points makes it difficult for the VLM to infer the correct navigation direction from the FPV, causing it to mistakenly interpret Frontier~1 as leading into a room. By contrast, Figure~\ref{fig:redpoint1} (right) shows that gaze guidance enables the VLM to correctly reason that Frontier~1 corresponds to a hallway rather than an adjacent room.

Overall, gaze points provide a focused spatial anchor for VLM analysis, replacing diffuse scene-level interpretation with specific spatial reasoning. Table~\ref{tab:alb_1} further quantifies the benefits of gaze-point guidance. Moreover, without gaze points, the VLM cannot reliably infer the corresponding focus region in the FPV from BEV frontier information alone, underscoring the critical role of gaze-guided spatial grounding.

\subsubsection{Unobserved Region Representation}

The representation of unobserved regions in FPVs has a substantial impact on VLM scene understanding. Empirically, using black or gray to denote unseen areas consistently leads to the failure cases shown in Figure~\ref{fig:black}. In particular, the VLM tends to misinterpret black regions as physical obstacles or hallucinate lighting-based spatial cues, resulting in unstable and unreliable predictions.

To mitigate this issue, we represent unobserved regions in purple in our experiments. This color is salient to VLMs while avoiding unintended associations with lighting variations or structural obstacles, thereby enabling more stable and accurate scene interpretation.

\section{Analysis of Re‑verification Cases}
\label{appendix:action}
\begin{figure*}[t]
    \centering
    \includegraphics[scale=0.7]{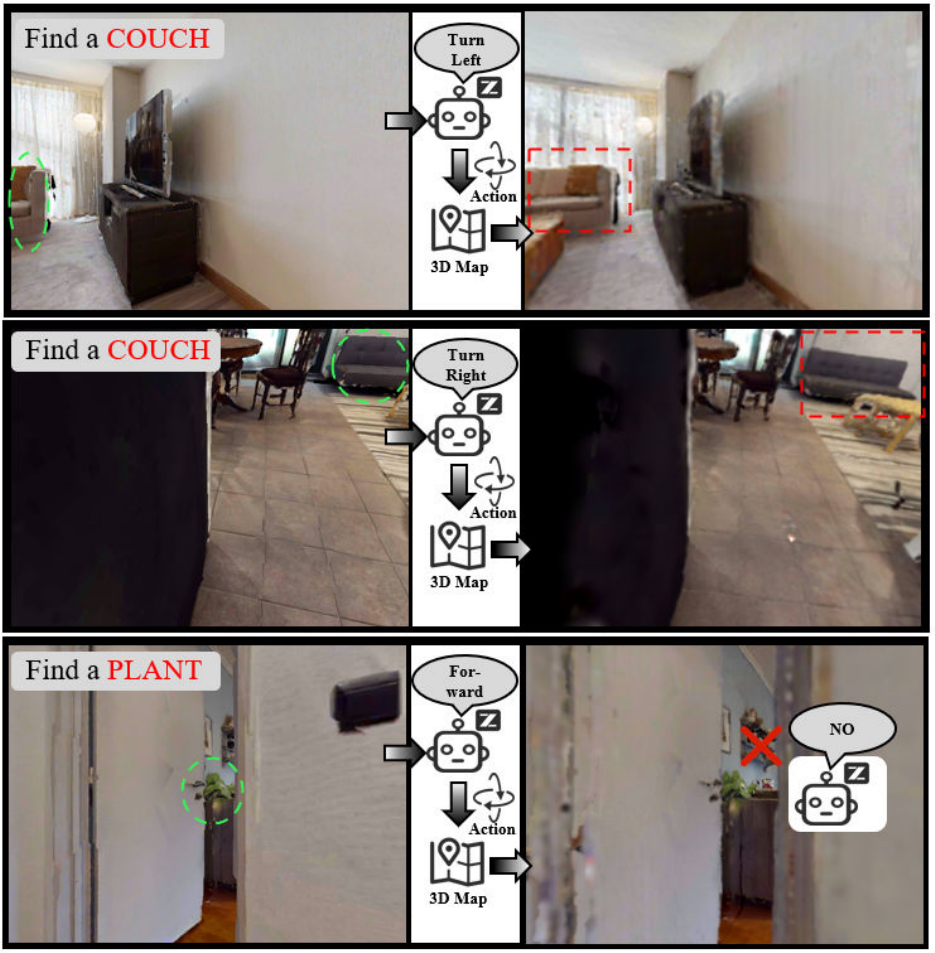}
    \caption{\textbf{Visualization of the action-decision VLM target re-verification.} The green circular bounding boxes indicate detections from the real-time detector, while the red bounding boxes denote the final detection results.}
    \label{fig:apaction}
\end{figure*}
\begin{figure*}[t]
    \centering
    \includegraphics[scale=0.45]{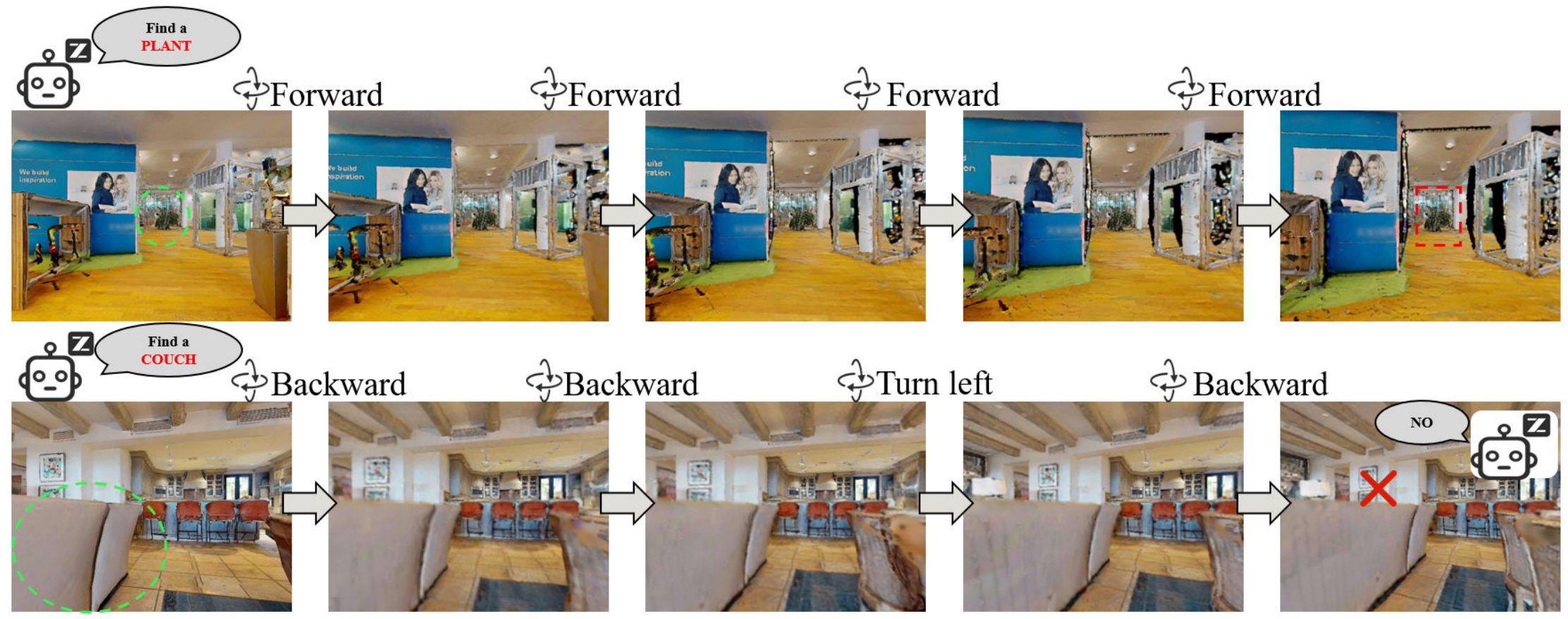}
    \caption{\textbf{Visualization of a long-horizon sequence of VLM-based action decisions for target re-verification.} The green circular bounding boxes indicate detections from the real-time detector, while the red bounding boxes denote the final detection results.}
    \label{fig:apaction1}
\end{figure*}
To prevent the action‑decision VLM from excessively influencing the agent’s behavior through overactive control, we impose an upper limit of five action decisions per verification cycle.

Figures~\ref{fig:apaction} and~\ref{fig:apaction1} illustrate the VLM‑driven active target re‑verification mechanism. Figure~\ref{fig:apaction} presents representative cases in which the VLM‑based action decision module performs target re‑verification and successfully corrects false positives generated by the real‑time detector. By reasoning over the target’s spatial location in the image, the VLM selects appropriate actions that induce viewpoint transformations, thereby expanding target visibility and reducing misdetections caused by limited local perception.

\section{Runtime Analysis}
\label{appendix:runtime}

\begin{table}[b]
    \centering
    \caption{\textbf{System runtime.} Average runtime of each component for a single run on an RTX 4060 Ti}
    \label{tab:runtime}
    \begin{tabular}{cccc}       
        \hline
        \multirow{2}{*}{Component} & \multicolumn{1}{c}{Free-viewpoint}  & \multicolumn{1}{c}{Guidance}  & \multicolumn{1}{c}{Frontier}   \\ 
                                    &optimization           &trajectory    & clustering\\ \hline
        Time(ms) & 115 & 2.4 & 6.3  \\ 
        \hline
    \end{tabular}
\end{table}
Table~\ref{tab:runtime} summarizes the runtime of the major system components. The cost of free‑viewpoint optimization is evaluated per optimization iteration, the guidance trajectory cost is computed per trajectory, and the frontier clustering cost corresponds to the total time required to group all frontier points.

\section{Real-World Experimental Details}
\label{appendix:realworld}
\subsection{Experiment Setup}
We employ the Go2 quadruped robot as the mobile platform. The system is powered by an NVIDIA Jetson AGX Orin embedded computing unit, featuring a 12‑core ARM v8.2 64‑bit CPU and 32GB of memory, running Ubuntu 20.04 and the Robot Operating System 2 (ROS2). The robot is equipped with an Intel RealSense D455 RGB‑D camera and a 6‑DoF robotic manipulator. The robot’s ego‑motion is estimated using legged odometry.

To mitigate motion‑induced jitter during locomotion, a pose‑following control strategy is applied at the manipulator end‑effector, allowing it to adaptively track the quadruped’s body posture in real time. The navigation policy is implemented within the Nav2 framework.
\begin{figure}[t]
    \centering
    \includegraphics[scale=0.5]{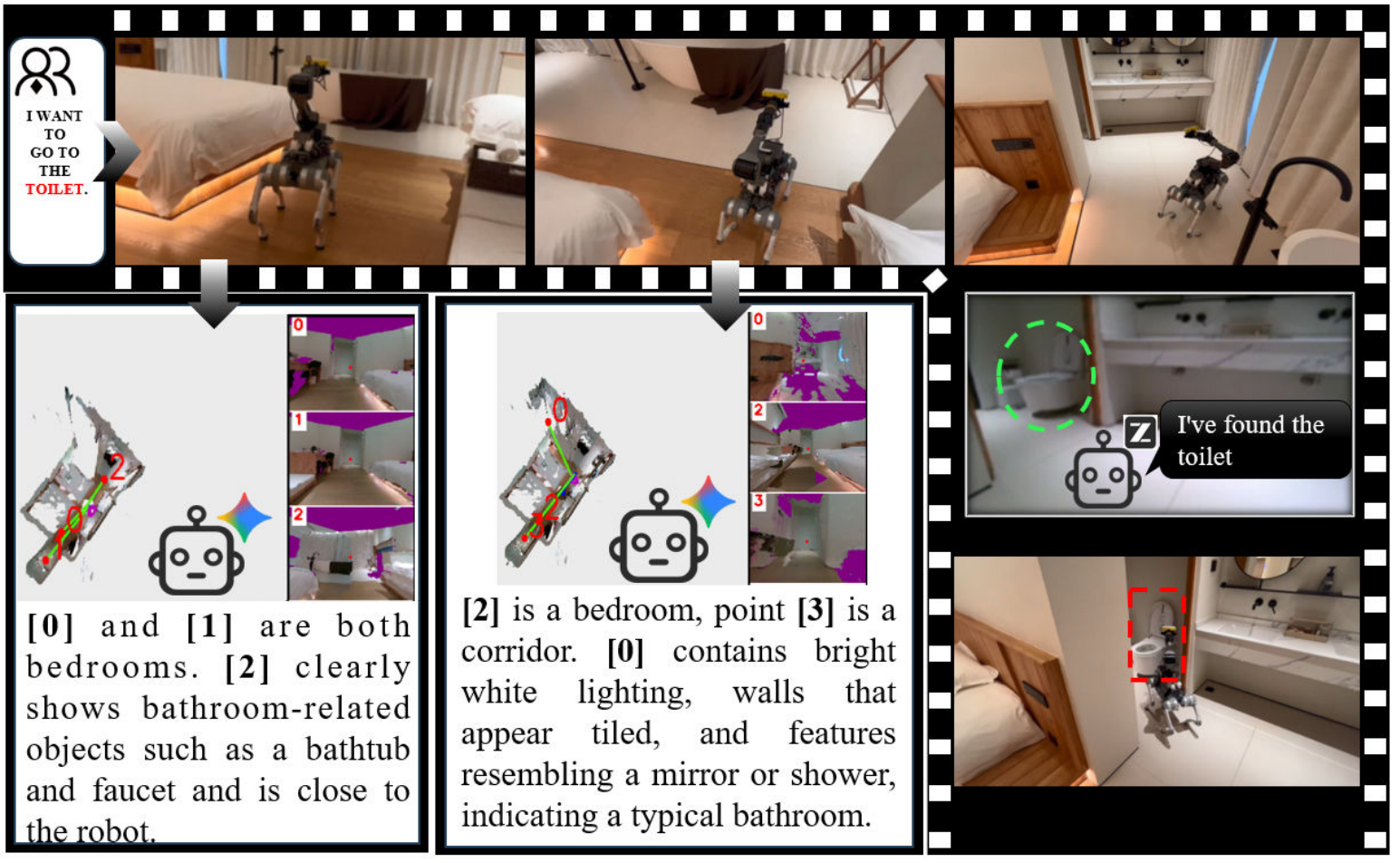}
    \caption{\textbf{3DGSNav Real-World Demonstration (Hard).} The robot successfully navigates to locate the toilet. The reasoning process is simplified in the figure.
    }
    \label{fig:real_toilet1}
\end{figure}
\begin{figure}[t]
    \centering
    \includegraphics[scale=0.5]{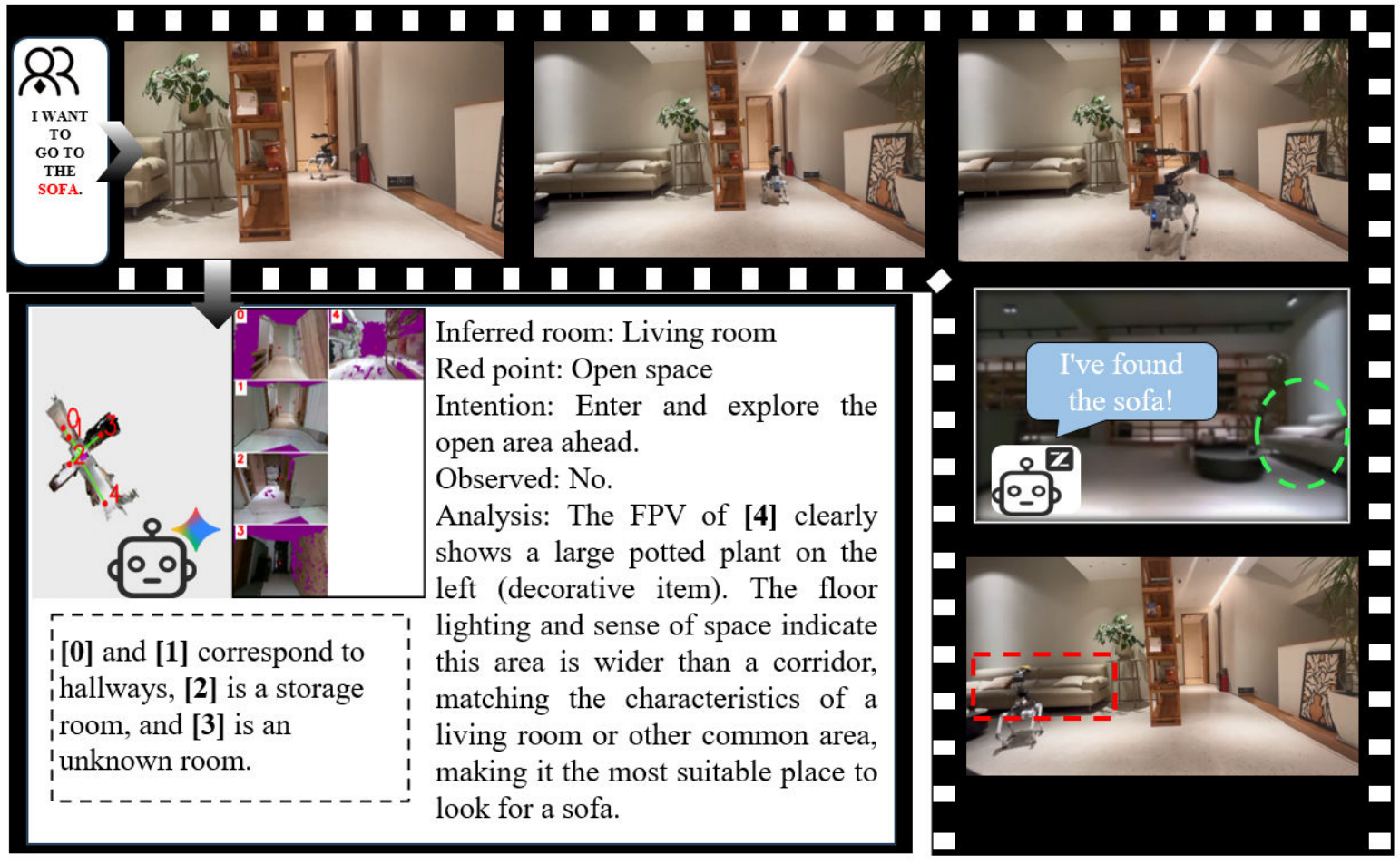}
    \caption{\textbf{3DGSNav Real-World Demonstration (Medium).} The robot successfully navigates to locate the sofa. The reasoning process is simplified in the figure.
    }
    \label{fig:real_sofa}
\end{figure}
\begin{figure}[t]
    \centering
    \includegraphics[scale=0.5]{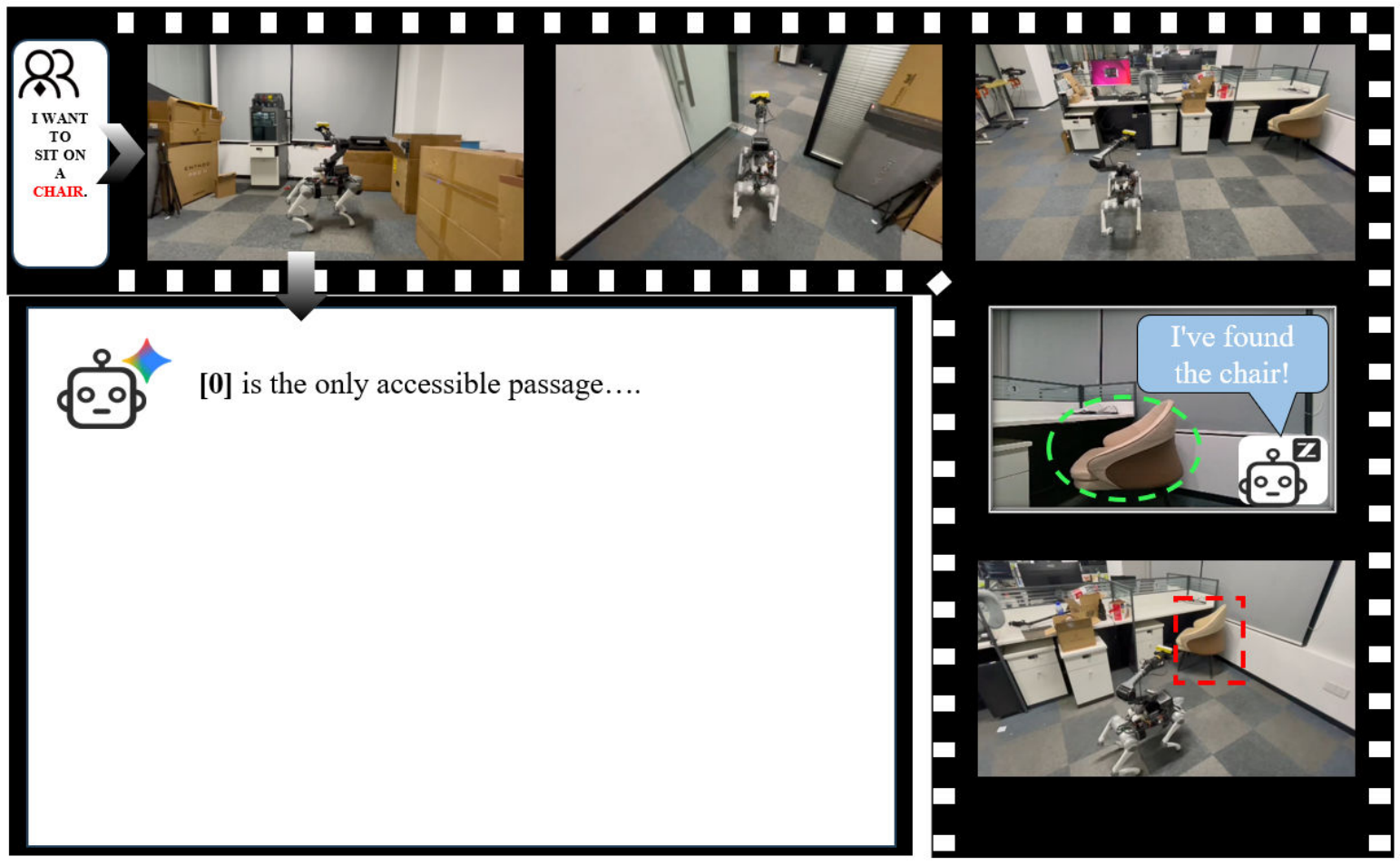}
    \caption{\textbf{3DGSNav Real-World Demonstration (Easy).} The robot successfully navigates to locate the chair. The reasoning process is simplified in the figure.
    }
    \label{fig:real_office_chair}
\end{figure}

\subsection{Real-World Experimental Results}
We design six groups of experiments with progressively increasing difficulty, as illustrated in Figures~\ref{fig:real_office_chair}, \ref{fig:real_sofa}, and~\ref{fig:real_toilet1}. Table~\ref{tab:realexp} presents the quantitative results, reporting the number of successful trials out of six runs for each experimental group. Overall, the results demonstrate the robustness and effectiveness of our proposed method in real‑world settings.

\begin{table}[b]
    \centering
    \vspace{-2mm}
    \caption{\textbf{Real‑World Experiments.} Failure cases include planning errors, collisions, and detection errors. }
    \label{tab:realexp}
    \begin{tabular}{c|c|ccc|cc}       
        \hline
          & \multicolumn{1}{c|}{Easy} & \multicolumn{3}{c|}{Medium} & \multicolumn{2}{c}{Hard} \\ \hline
        Scenario & Office & Hotel & Hotel & Hotel & Hotel & Hotel \\
        Target   & chair  & sofa  & plant & chair & toilet & slippers \\ \hline
        Success/Total & 5/6 & 4/6 & 5/6 & 4/6 & 4/6 & 3/6 \\
        \hline
    \end{tabular}
\end{table}

\subsection{Real-World Experimental Analysis}
\subsubsection{Motion Control and System Stability Issues} 
First, the quadruped robot’s native locomotion policy is not optimized to handle the additional payload introduced by the manipulator. This results in a shifted center of mass, which in turn causes deviations in the walking trajectory. Second, severe vibrations during locomotion lead to motion blur in the visual input. Although a posture‑following strategy is employed to stabilize the camera, its effectiveness remains limited under dynamic motion. Furthermore, the high latency of the VLM API forces the motors to maintain static postures for prolonged periods, which can cause overheating and eventually trigger system protection mechanisms.

To address these challenges, future work will focus on tightly integrating manipulator‑based active perception with the quadruped’s locomotion policy. This integration aims to achieve unified and coordinated control over perception and motion, thereby mitigating payload imbalance, improving stability, and reducing the risk of overheating.

\subsubsection{Visual Perception Limitations}
Multiple experiments suggest that the RGB-D camera is prone to perception failures when encountering glass surfaces, strong specular reflections, and highly hollow or open-structured environments. These failures degrade the rendering quality of FPVs, thereby impairing the reasoning performance of the VLM and compromising navigation safety.

\section{Chain-of-Thought prompt Design}
\label{appendix:cot}
\begin{figure}[t]
    \centering
    \includegraphics[scale=0.6]{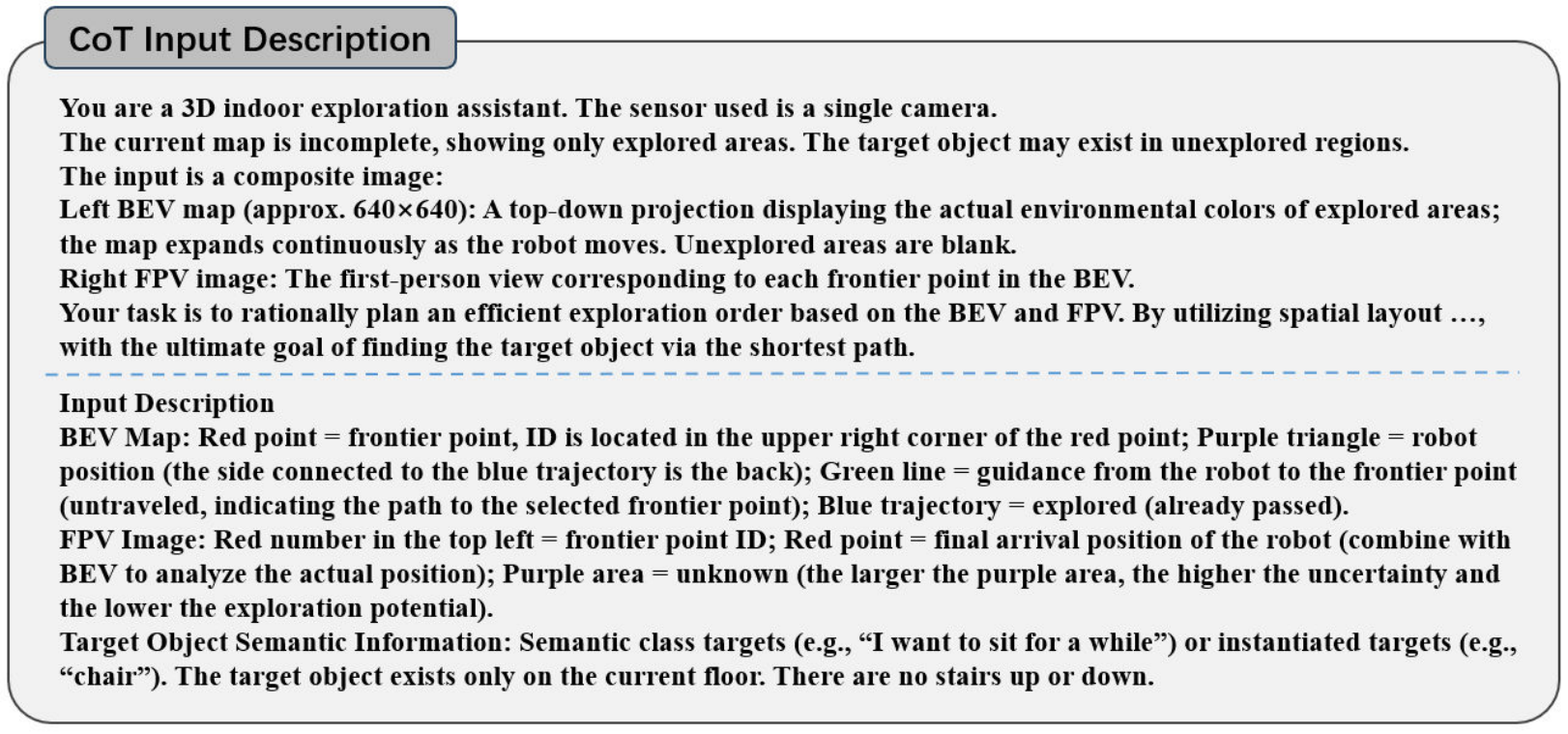}
    \caption{\textbf{CoT input description.}
    }
    \label{fig:cotinput}
\end{figure}

\begin{figure}[t]
    \centering
    \includegraphics[scale=0.55]{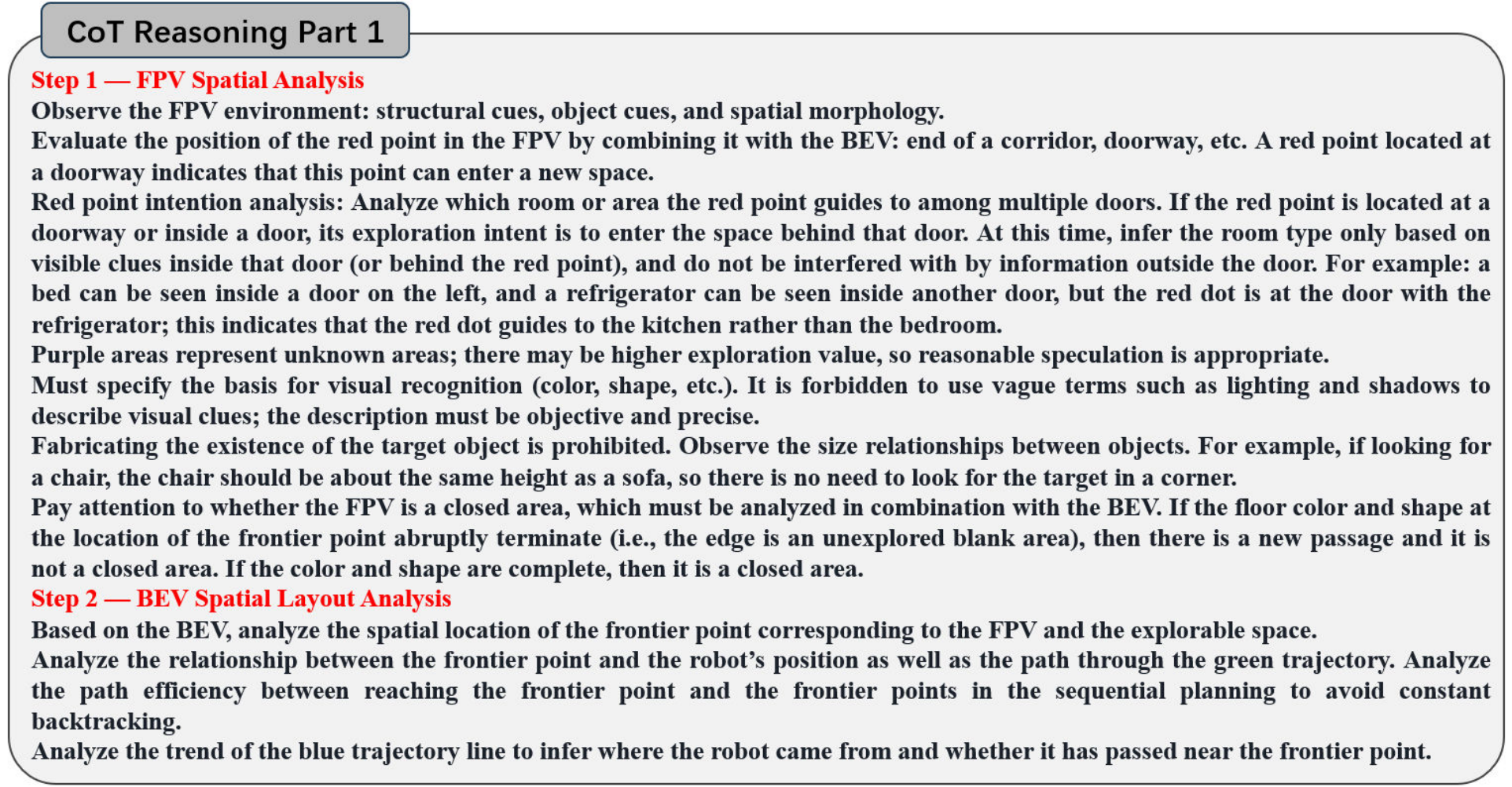}
    \caption{\textbf{CoT reasoning part 1.}
    }
    \label{fig:cotreason1}
\end{figure}

\begin{figure}[t]
    \centering
    \includegraphics[scale=0.48]{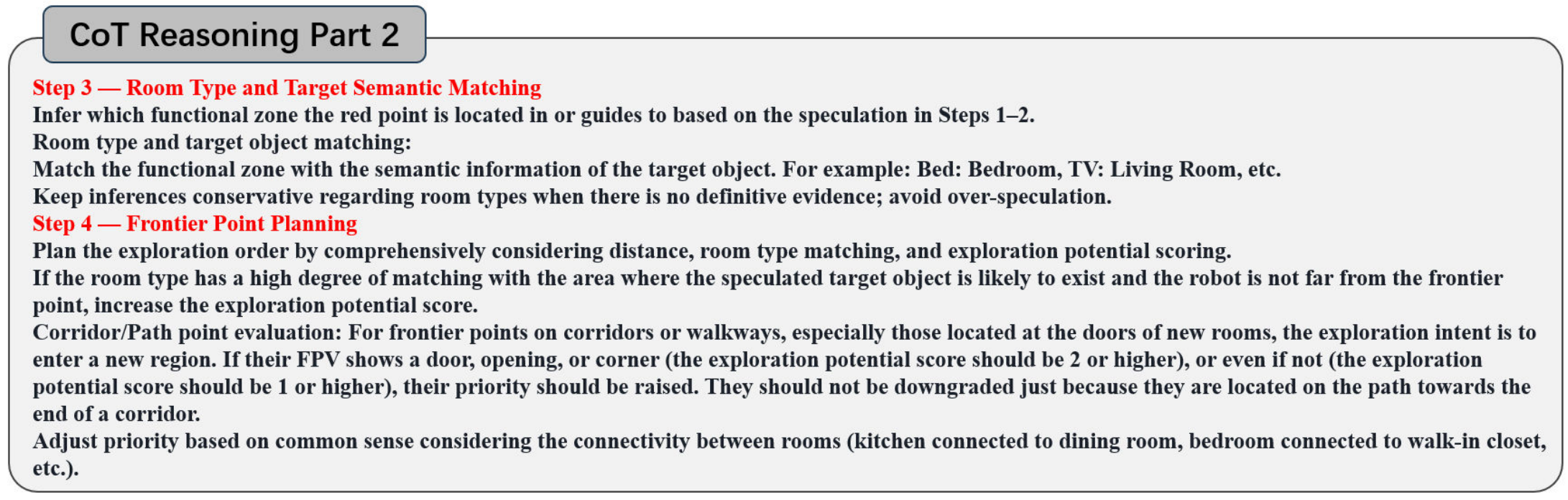}
    \caption{\textbf{CoT reasoning part 2.}
    }
    \label{fig:cotreason2}
\end{figure}

\begin{figure}[t]
    \centering
    \includegraphics[scale=0.5]{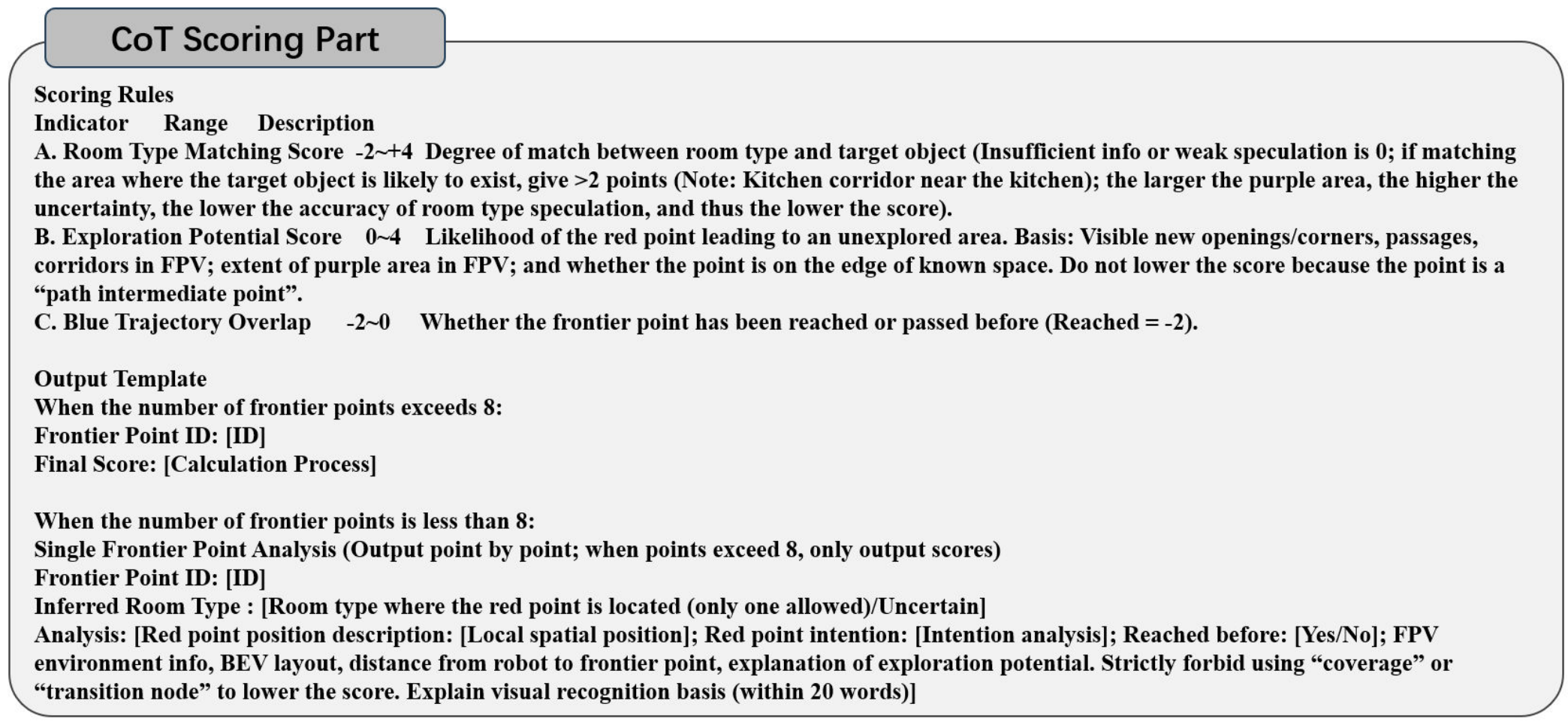}
    \caption{\textbf{CoT scoring part.} 
    }
    \label{fig:cotscore}
\end{figure}

\begin{figure}[t]
    \centering
    \includegraphics[scale=0.6]{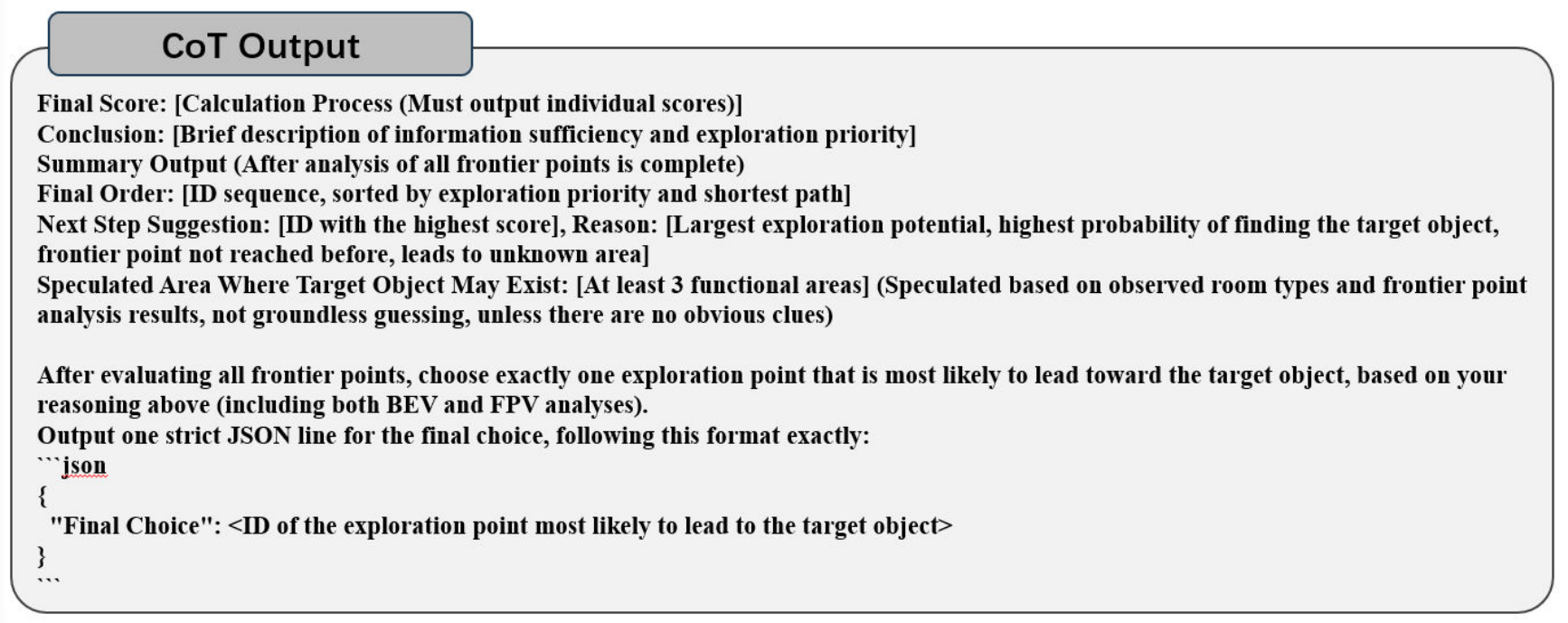}
    \caption{\textbf{CoT output.} 
    }
    \label{fig:cotoutput}
\end{figure}

Figures~\ref{fig:cotinput}, \ref{fig:cotreason1}, \ref{fig:cotreason2}, \ref{fig:cotscore}, and~\ref{fig:cotoutput} illustrate the key components of the CoT prompts used in our approach, including the input formulation, intermediate reasoning steps, scoring mechanism, and final output generation.




\end{document}